\documentclass{article}

\PassOptionsToPackage{numbers, compress}{natbib}
 \usepackage[preprint]{neurips_2026}

\usepackage[utf8]{inputenc} 
\usepackage[T1]{fontenc}    
\usepackage{url}            
\usepackage{amsfonts}       
\usepackage{nicefrac}       
\usepackage{microtype}      

\usepackage{hyperref}
\usepackage{graphicx}
\usepackage{amsthm}
\usepackage[ruled,linesnumbered,vlined,noend]{algorithm2e}
\usepackage{color}
\usepackage{tikz}
\usepackage{tikz-qtree}
\usepackage{pgfplots}
\usepackage{pgfplotstable}
\usepackage[hypcap=true]{subcaption}
\usetikzlibrary{patterns}
\usepackage{tabularx}
\usepackage{makecell}
\usepackage{multirow}
\usepackage{booktabs}
\usepackage{bm}
\usepackage{balance}
\usepackage{stfloats}
\usepackage[table,xcdraw]{xcolor}
\usepackage{caption} 
\usepackage{enumitem}

\usepackage{utfsym} 

\definecolor{linecolor}{rgb}{0.55, 0, 0} 
\definecolor{bgcolor}{rgb}{1, 0.95, 0.95} 

\definecolor{best}{HTML}{F8B2A2}    
\definecolor{second}{HTML}{F6DFD6}

\newcommand{\Yes}{{\color{green} \usym{2713}}}
\newcommand{\No}{{\color{red} \usym{2717}}}
\usepackage[most]{tcolorbox}

\tcbset{highlightcode/.style=
    {enhanced, 
    colframe=gray!20, 
    colback=gray!20, 
    boxrule=0pt, 
    boxsep=0pt, 
    left=1mm, 
    right=0mm, 
    top=1mm, 
    bottom=1mm,
    overlay={
        \begin{scope}[shift={(frame.north west)}]
            \foreach \i in {1,2,3} {
                \node[anchor=east] at (-5mm,{-\i*15mm + 1.5mm}) {\footnotesize\the\numexpr\value{AlgoLine}+\i\relax};
            }
        \end{scope}
    },
}}

\definecolor{mycolor2}{RGB}{164,224,187} 
\definecolor{mycolor1}{RGB}{53,122,162} 

\newcommand{\colorcell}[1]{%
    \cellcolor{mycolor1!#1!mycolor2}%
    \ifnum#1>49%
        {\huge \textcolor{white}{{#1}}}
    \else
        {\huge \textcolor{black}{{#1}}}
    \fi
}

\usepackage{colortbl}
\usepackage{etoolbox}

\newcommand{\hhcell}[1]{%
    \cellcolor{mycolor1!#1!mycolor2}%
    \ifnum#1>49%
        {\huge\textcolor{white}{#1}}%
    \else
        {\huge\textcolor{black}{#1}}%
    \fi
}

\usepackage{float}
\usepackage{xspace}
\tcbset{
  aibox/.style={
    width=474.18663pt,
    top=10pt,
    colback=white,
    colframe=black,
    colbacktitle=black,
    enhanced,
    center,
    attach boxed title to top left={yshift=-0.1in,xshift=0.15in},
    boxed title style={boxrule=0pt,colframe=white,},
  }
}
\newtcolorbox{AIbox}[2][]{aibox,title=#2,#1}

\usepackage{pifont}

\title{ASTRA-QA: A Benchmark for Abstract Question Answering over Documents}

\author{%
  Shu Wang$^{1}$ ~Shansong Zhou$^{1}$ ~Xinyang Wang$^{1}$ ~Shiwei Wang$^{2}$ ~Hulong Wu$^{2}$ ~Yixiang Fang$^{1}$\thanks{Corresponding author.}\\
  $^{1}$The Chinese University of Hong Kong, Shenzhen\\
  $^{2}$Data Science Group, Huolala\\
  \texttt{shuwang3@link.cuhk.edu.cn, fangyixiang@cuhk.edu.cn}\\
}

\usepackage{listings}

\lstdefinestyle{promptstyle}{
  basicstyle=\rmfamily\fontsize{8.55}{10.15}\selectfont,
  columns=fullflexible,
  keepspaces=true,
  showstringspaces=false,
  breaklines=true,
  breakatwhitespace=false,
  backgroundcolor=\color{black!2},
  frame=none,
  xleftmargin=0pt,
  xrightmargin=0pt,
  aboveskip=0pt,
  belowskip=0pt,
  numbers=none,
  captionpos=b
}
\lstnewenvironment{promptlisting}[1][]{\lstset{style=promptstyle,#1}}{}
\newtcblisting{PromptBox}[2][]{%
  ragstyle,
  enhanced,
  listing only,
  colback=black!2,
  colframe=black!75,
  boxrule=0.65pt,
  toprule=1.4pt,
  bottomrule=1.0pt,
  left=1.6mm,
  right=1.6mm,
  top=1.0mm,
  bottom=1.0mm,
  colbacktitle=black,
  coltitle=white,
  fonttitle=\bfseries\sffamily\fontsize{7.8}{8.6}\selectfont,
  title={#2},
  before skip=0.35em,
  after skip=0.3em,
  listing options={style=promptstyle},
  #1
}

\tcbset{
  ragstyle/.style={
    colback=white,
    colframe=black,
    boxrule=0.8pt,
    toprule=1.5pt,
    bottomrule=1.5pt,
    arc=0mm,
    left=2mm,
    right=2mm,
    top=2mm,
    bottom=2mm,
    width=\textwidth,
    enlarge left by=0mm,
    before skip=1em,
    after skip=1em
  }
}

\newcommand{\modelname}{ASTRA-QA}

\begin{document}

\maketitle

\begin{abstract}
Document-based question answering (QA) increasingly includes abstract questions that require synthesizing scattered information from long documents or across multiple documents into coherent answers.
However, this setting is still poorly supported by existing benchmarks and evaluation methods, which often lack stable abstract references or rely on coarse similarity metrics and unstable head-to-head comparisons.
To alleviate this issue, we introduce {\bf \modelname}, a benchmark for \textbf{A}b\textbf{STRAeru}ct \textbf{Q}uestion \textbf{A}nswering over documents.
{\modelname} contains 869 QA instances over academic papers and news documents, covering five abstract question types and three controlled retrieval scopes.
Each instance is equipped with explicit evaluation annotations, including answer topic sets, curated unsupported topics, and aligned evidence.
Building on these annotations, {\modelname} assesses whether answers cover required key points and avoid unsupported content by directly scoring topic coverage and curated unsupported content, enabling scalable evaluation without exhaustive head-to-head comparisons.
Experiments with representative Retrieval-Augmented Generation (RAG) methods spanning vanilla, graph-based, and hierarchical retrieval settings show that {\modelname} provides reference-grounded diagnostics for coverage, hallucination, and retrieval-scope robustness.
Our dataset and code are available at \url{https://xinyangsally.github.io/astra-benchmark}.
\end{abstract}


\section{Introduction}
\label{sec:intro}

Retrieval-augmented generation (RAG) has been increasingly used for improving the faithfulness of large language models (LLMs) in question answering (QA) by grounding generation in external documents~\cite{lewis2020retrieval}.
%
%
Recently, the abstract questions~\cite{edge2024local,guo2024lightrag,wang2026archrag,zhou2025depth,chen2026pathrag,huang2025retrieval} in the RAG setting have received plenty of attention, and they are prevalent in practical scenarios, including long-form QA~\cite{cai2024forag}, query-focused summarization~\cite{edge2024local}, and scientific document analysis~\cite{dasigi2021dataset,baumgartner2025peerqa}.
For example, a user may ask {\it ``What are the main differences between GraphRAG and LightRAG, and what are the strengths and limitations of each method?''}
Compared with specific fact-seeking questions, they are high-level questions requiring document-level synthesis rather than isolated fact retrieval.
As a result, answering an abstract question requires the RAG system to synthesize information over a broad document scope and produce a coherent, selective, and faithful response.


Existing benchmarks have advanced long-form QA, scientific document QA, multi-hop reasoning, long-context understanding, and factual RAG evaluation, as seen in ASQA~\cite{stelmakh2022asqa}, Qasper~\cite{dasigi2021dataset}, PeerQA~\cite{baumgartner2025peerqa}, HotpotQA~\cite{yang2018hotpotqa}, LongBench~\cite{bai2024longbench}, and CRAG~\cite{yang2024crag}.
However, these benchmarks are designed around different primary goals, such as ambiguous factoid resolution, local evidence lookup, reasoning chains, general long-context understanding, and factual retrieval.
As a result, they neither directly target synthesis-heavy abstract questions that require methods to integrate information across broad document contexts, nor do they provide explicit evaluation annotations for fine-grained diagnostic evaluation.
What remains inadequately explored is not merely the retrieval of more accurate information, but rather the ability to perform abstract synthesis based on retrieved evidence under controlled evaluation settings.

\begin{table}[t]
\centering
\caption{Comparison between {\modelname} and representative QA benchmarks. Hall-Eval and Multi-Scope denote hallucination evaluation and multiple retrieval scopes, respectively.}
\label{tab:benchmark-comparison}
\small
\setlength{\tabcolsep}{2pt}
\begin{tabular}{lccccc}
\toprule
{\bf Benchmark} & {\bf Target} & {\bf Evaluation} & {\bf Answer Topics} & {\bf Hall-Eval} & {\bf Multi-Scope} \\
\midrule
PeerQA~\cite{baumgartner2025peerqa} & Scientific QA & LLM judge & \No & \No & \No \\
HotpotQA~\cite{yang2018hotpotqa} & Multi-hop QA & EM/F1 & \No & \No & \No \\
LongBench~\cite{bai2024longbench} & Long-context & Mixed metrics & \No & \No & \No \\
CRAG~\cite{yang2024crag} & Factual RAG & LLM judge & \No & \Yes & \Yes \\
RAG-QA Arena~\cite{han2024rag} & Long-form RAG & Head-to-head comparison & \No & \No & \No \\
LiveRAG~\cite{carmel2025liverag} & Diverse QA & Auto. + LLM metrics & \No & \No & \Yes \\
\hline
{\bf {\modelname} (Ours)} & Abstract QA & Topic-based scoring & \Yes & \Yes & \Yes \\
\bottomrule
\end{tabular}
\end{table}

This mismatch also poses an evaluation challenge for abstract QA.
Conventional lexical and semantic metrics, such as ROUGE~\cite{lin2004rouge} and BERTScore~\cite{zhang2019bertscore}, are too coarse-grained for long-form abstract answers, whose wording may vary substantially while differing in topical coverage.
QA-based and factuality-oriented metrics, such as QuestEval~\cite{scialom2021questeval}, QAFactEval~\cite{fabbri2022qafacteval}, and FActScore~\cite{min2023factscore}, improve grounding analysis, but they mainly focus on consistency or factual precision rather than whether a response covers all required answer topics.
LLM-based evaluators, such as G-Eval~\cite{liu2023g} and head-to-head comparison~\cite{edge2024local,guo2024lightrag,wang2026archrag}, can provide useful holistic judgments, but they are costly to scale and do not explicitly ground evaluation in stable answers or source evidence.
A benchmark for abstract QA, therefore, requires explicit evaluation annotations that support direct and fine-grained measurement of topic coverage and reference-targeted unsupported content. 
Table~\ref{tab:benchmark-comparison} summarizes the key characteristics of {\modelname} and representative QA benchmarks.

\begin{figure}[h]
    \centering
    \setlength{\abovecaptionskip}{0.1cm}
    \setlength{\belowcaptionskip}{-0.2cm}
    \includegraphics[width=0.95\linewidth,trim=0 2 0 0,clip]{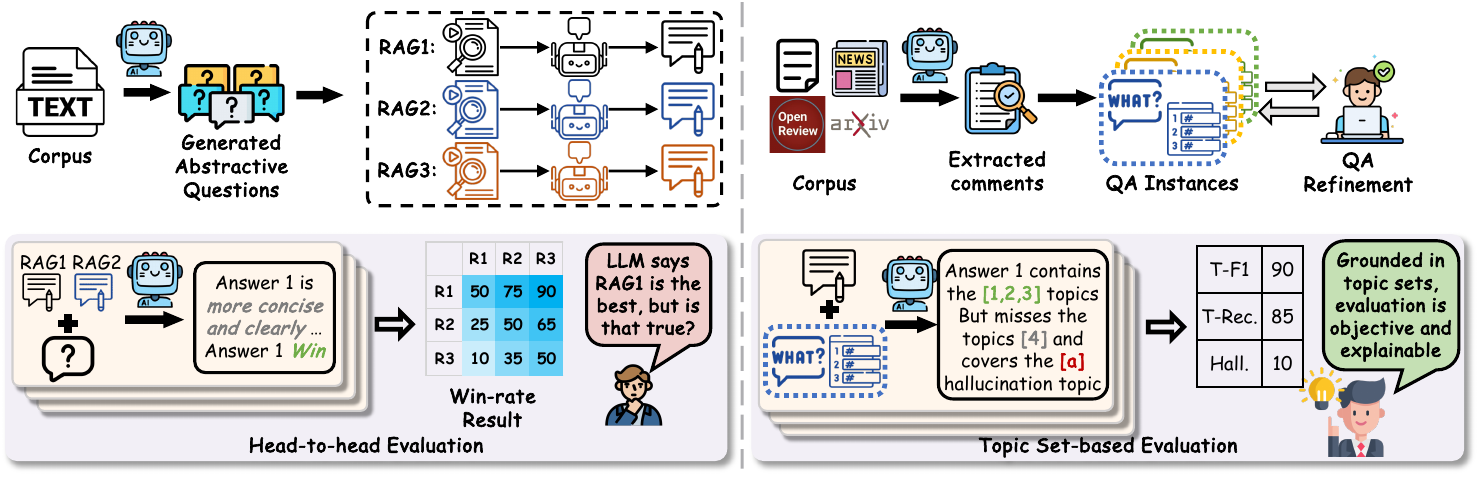}
    \caption{Comparison between existing head-to-head evaluation and our topic-based evaluation for abstract QA.}
    \label{fig:alpha}
\end{figure}

To address these gaps, we introduce \textbf{\modelname}, a benchmark for \textbf{AbSTRA}ct \textbf{QA} over documents, as shown in Figure~\ref{fig:alpha}.
%
%
Its design follows a simple grading intuition: a good answer should cover the required key points while avoiding unsupported content.
Accordingly, each {\modelname} instance is annotated with a set of answer topics, a curated hallucination set, aligned evidence, and retrieval-scope metadata.
{\modelname} contains 869 QA instances over more than 700 paper documents and 1,500 news articles, covering five synthesis-heavy question types and three retrieval scopes.
Built on these references, our evaluation method directly scores topic-based coverage and curated hallucination-topic matches.

In summary, our principal contributions are as follows:
\begin{itemize}[nosep,leftmargin=1.2em,labelsep=0.4em]

\item We introduce {\modelname}, a new benchmark for abstract QA over academic papers and news documents, covering single-document summarization, two-way comparison, multi-way comparison, thematic enumeration, and temporal reasoning.

\item We propose a topic-based evaluation method that directly scores topic coverage and hallucinated content, avoiding costly head-to-head judgments.

\item We evaluate representative RAG methods on {\modelname} and analyze their behavior across five question types and three retrieval scopes, highlighting the benchmark's ability to surface coverage, hallucination, and retrieval-scope challenges.
\end{itemize}
\section{Related Work}
\label{sec:related-work}

RAG has been widely used for knowledge-intensive QA, where retrieved documents help LLMs produce factual answers beyond parametric memory alone~\cite{lewis2020retrieval,izacard2021leveraging}.
Recent RAG methods~\cite{edge2024local,guo2024lightrag,wang2026archrag,huang2025retrieval,gutierrez2024hipporag,sarthi2024raptor} have moved further toward long-form and document-level synthesis through hierarchical retrieval, graph-structured indexing, and long-context reading.
These methods show that summarization, comparison, and broader corpus-level synthesis have become realistic RAG targets.
However, abstract questions still lack dedicated datasets and an evaluation method to assess whether generated answers are complete, grounded, and free of unsupported additions.


\textbf{Benchmarks for RAG.}
Existing QA benchmarks for RAG include open-domain retrieval, multi-hop reasoning, scientific document QA, and long-context understanding, as seen in ELI5~\cite{fan2019eli5}, ASQA~\cite{stelmakh2022asqa}, HotpotQA~\cite{yang2018hotpotqa}, Qasper~\cite{dasigi2021dataset}, PeerQA~\cite{baumgartner2025peerqa}, and LongBench~\cite{bai2024longbench}. 
These resources establish the importance of long answers, evidence supervision, and difficult retrieval settings, but most still center on factoid resolution, local information seeking, or general long-context evaluation.
They do not jointly provide curated abstract references for comparison, enumeration, and temporal synthesis, nor do they expose topic-level targets that make answer completeness and hallucination behavior easy to diagnose.
{\modelname} is designed for this missing setting by providing curated QA instances with reference topic sets, hallucination sets, and aligned evidence spans over academic papers and news documents.

\textbf{Evaluation for Abstract QA.}
Early evaluation of answers to abstract questions relied on lexical or semantic reference matching, such as ROUGE~\cite{lin2004rouge}, BERTScore~\cite{zhang2019bertscore}, and QA-based or factuality metrics such as FEQA~\cite{durmus2020feqa}, QuestEval~\cite{scialom2021questeval}, QAFactEval~\cite{fabbri2022qafacteval}, and FActScore~\cite{min2023factscore}.
These metrics provide useful surface, semantic, or grounding signals, but they do not jointly diagnose whether a response covers all required topics and avoids unsupported additions.
With the development of LLM judges, reference-guided evaluation offers another direction~\cite{zheng2023judging}, as in RAG-QA Arena~\cite{han2024rag} and G-Eval~\cite{liu2023g}.
Head-to-head comparison is also widely used in abstract RAG evaluation, where an LLM judge directly compares two RAG system outputs~\cite{edge2024local,guo2024lightrag,wang2026archrag,chen2026pathrag,huang2025retrieval}.
Such evaluations are useful for holistic preference judgments, but they make it difficult to compare systems under stable reference targets or to obtain diagnostic coverage and hallucination metrics.
{\modelname} therefore emphasizes interpretable reference targets for coverage and hallucination analysis rather than relying only on answer-level similarity or pairwise preference.


\section{Our {\modelname} Benchmark}
\label{sec:dataset}

In this section, we present the construction process of {\modelname}, including its design goals, construction pipeline, question types, and dataset statistics.

\subsection{Overview of {\modelname}}
\label{sec:overview}

{\modelname} is a benchmark for abstract QA over documents, with a focus on evaluating RAG methods.
Unlike conventional QA benchmarks that emphasize short answers or extractive evidence lookup, {\modelname} is designed to evaluate whether a RAG system can synthesize information from long documents and produce responses that are coherent, selective, and faithful to the source content.
Such questions commonly require document-level summaries, structured comparisons, thematic organization, and temporally grounded synthesis rather than isolated factual snippets.
{\modelname} is built over two source domains, namely academic papers and news documents, and covers five abstract question types.

\begin{figure}[t]
    \centering
    \setlength{\abovecaptionskip}{0.1cm}
    \setlength{\belowcaptionskip}{-0.2cm}
    \includegraphics[width=0.95\linewidth,trim=0 0 0 0,clip]{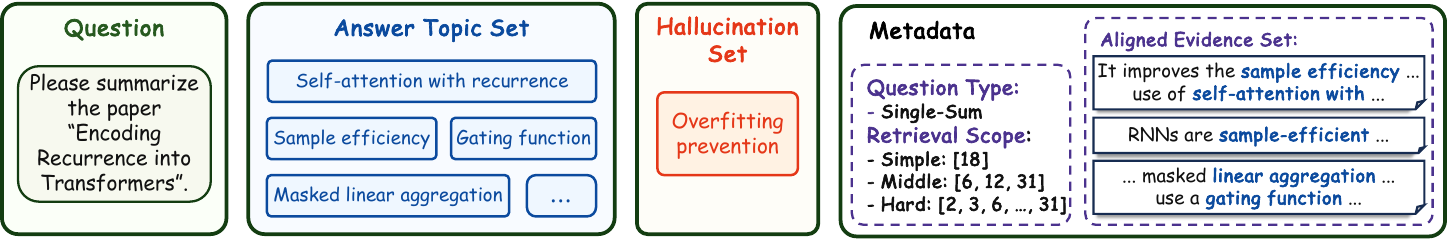}
    \caption{An example QA instance in the {\modelname} dataset.}
    \label{fig:qa}
\end{figure}

Formally, {\modelname} consists of a document corpus $D$ and a set of QA instances.
Each QA instance is represented as $(Q, A, H, M)$, where $Q$ is an abstract question, $A = \{\tau_1, \tau_2, \cdots, \tau_n\}$ is the answer topic set, $H = \{h_1, h_2, \cdots, h_k\}$ is a curated hallucination set containing plausible but unsupported topics, and $M$ denotes the associated metadata, including the aligned evidence set $E$, the question type, and the retrieval scope.
%
Figure~\ref{fig:qa} provides an example of this representation, showing the question, answer topic set, hallucination set, and metadata of a QA instance.
The set $A$ is constructed from high-level summary signals and used by our topic-based evaluation method to assess answer coverage, while $H$ records relevant but unsupported topics and is used to assess whether responses include these curated hallucination targets.

\subsection{Construction Pipeline}

\begin{figure}[h]
    \centering
    \setlength{\abovecaptionskip}{0.1cm}
    \includegraphics[width=0.95\linewidth,trim=12 6 8 0,clip]{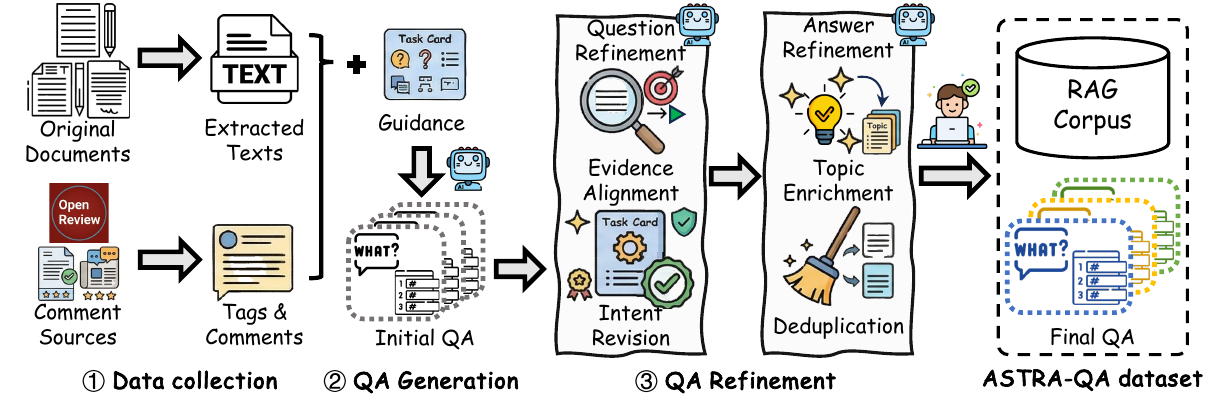}
    \caption{Workflow for constructing the {\modelname} dataset.}
    \label{fig:workflow}
\end{figure}

{\modelname} is constructed through a three-stage construction pipeline, as illustrated in Figure~\ref{fig:workflow}.
The pipeline converts heterogeneous source materials into abstract QA instances grounded in curated document collections, with reference-grounded evidence and comprehensive topic-set answers.

\textbf{Step 1: Data collection.} 
Original documents are collected from OpenReview, focusing on ICLR 2023~\cite{soergel2013openreview}, survey papers from arXiv\footnote{\url{https://arxiv.org/}}, the tagged corpus of publications of Epstein et al.~\cite{epstein2020mapping}, and news articles downloaded through the mediastack API\footnote{\url{https://mediastack.com/}}, forming an initial candidate pool of 700+ paper documents and 1,500+ news articles.
4,700+ comments are gathered separately from OpenReview reviews\footnote{\url{https://openreview.net/}} for ICLR papers (including reviewer comments: summary, strengths, and weaknesses), tags from corpus of publications of Epstein et al.~\cite{epstein2020mapping}, and metadata associated with the mediastack news records.
We then perform \emph{content extraction} on both the original documents and the matched comments.
Specifically, for PDF documents, we use MinerU~\cite{wang2024mineru} to parse the document text, whereas for HTML documents, we extract the page content directly.
We finally align each comment with its corresponding source document and merge all documents into a corpus $D$.

\textbf{Step 2: QA instance generation.} 
We use the processed materials, along with \emph{type-specific guidance}, to generate initial QA instances with an LLM.
Specifically, we use GPT-4o~\cite{hurst2024gpt} throughout the {\modelname} construction pipeline for generation and refinement.
The LLM input includes the type guidance, the matched comment text, the content of the original documents, and a small number of examples.
Each question $q$ is required to match the question type, and its answer $A$ is represented as a \textit{topic set} extracted from matched comments, tags, and other summary-like context aligned to the source documents, with 17.03 topics on average.
We design five common abstract question types, including summarization, pair comparison, multi-target comparison, enumeration, and temporal reasoning, and instantiate a different prompt template for each type.
The type-specific generation details are described in Section~\ref{sec:adc-tasks-annotation}.

\textbf{Step 3: QA instance refinement.} 
We refine the generated question and the topic-set answer separately.
{\bf (1)~Question refinement}. We perform {\it evidence alignment} by checking whether the question can be answered from the selected evidence documents alone.
To do so, we retrieve 10 related but non-evidence documents as probes and use an LLM to judge whether the question depends on information outside the designated evidence scope.
We then revise the \textit{question intent} and enforce consistency with the assigned question type, e.g., by making the comparison dimension explicit.
{\bf (2)~Answer refinement}. We first perform topic enrichment.
We retrieve 5 similar questions to obtain candidate supplemental topics.
We then use an LLM to judge, based on the original question, the original evidence, and the matched comment, whether each candidate topic should be included in $A$.
Accepted topics are added to $A$, while rejected but plausible unsupported topics are recorded in the curated \emph{hallucination set} $H$ for later evaluation.
We next {\it deduplicate} the enriched topic set $A$ by using an LLM to merge or remove repeated and semantically equivalent topics.
This stage turns model-generated drafts into curated benchmark samples.
After manual review, we {\it filter out} low-quality QA instances and their associated documents, resulting in the final {\modelname} dataset.

\subsection{Question Type Design and Answer Annotation}
\label{sec:adc-tasks-annotation}

We next describe the five common types of abstract questions covered by {\modelname} and explain how the corresponding QA instances are constructed.
Example QA instances for five question types are provided in the appendix.

\begin{itemize}[nosep,leftmargin=0.8em]
\item \textbf{\textsc{Single-Sum} (single-document summarization).} Single-Sum asks for a summary of one paper or one document-centered source, targeting the overall content of the document rather than a local fact or passage-level detail.
We construct this question type by providing the LLM with the original document, reviewer comments, and examples, and then generating a summary question together with a topic-set answer.

\item \textbf{\textsc{Pair-Comp} (two-way comparison).} Pair-Comp is used when the source text presents two methods, entities, documents, or events under an explicit contrast.
We use survey passages or other summary-like evidence that explicitly compare two targets, and require the answer to be organized into two separate target-specific topic sets rather than a merged summary.

\item \textbf{\textsc{Multi-Comp} (multi-way comparison).} Multi-Comp extends comparison to settings with at least three items.
Its construction relies on survey-style comparison passages or other summary-like text that place multiple targets under a shared discussion scope, and the answer is required to keep a consistent target-by-target structure across all compared items.

\item \textbf{\textsc{Enumeration} (thematic enumeration).} Enumeration asks for as many relevant items as possible under a shared theme, such as contributions, reasons, or findings.
We construct this question from summary-like passages that discuss multiple aspects of the same topic, and require the answer to be organized as a list of parallel topic items with minimal redundancy.

\item \textbf{\textsc{Temporal} (temporal reasoning).} 
Temporal asks about an event, fact, or state within a specified time range, or about how it changes over time.
Its construction relies on time-bounded news documents and related summary-like text, and the answer is required to reflect temporal stages or updates rather than a static summary.

\end{itemize}

To evaluate how the retrieval range affects RAG performance on abstract questions, we define three retrieval settings, denoted as \textsc{Simple}, \textsc{Middle}, and \textsc{Hard}, by varying the retrieval span while keeping the reference answer fixed.
\textsc{Simple}: the search space is restricted to the exact document or tightly scoped documents that directly support the question; for single-document summarization, this reduces to the source document itself.
\textsc{Middle}: we cluster semantically related questions and merge the evidence documents within each cluster, yielding a broader but still relevant corpus.
\textsc{Hard}: all documents belonging to the same question type are pooled together, creating the largest retrieval space and introducing substantially more distractors.
This design allows {\modelname} to probe how RAG methods behave as the search space expands from the most precise corpus scope to broader and more noise-heavy corpora.

\subsection{Dataset Statistics}
\label{sec:adc-statistics}

\begin{figure}[t]
\centering

\begin{minipage}[t]{0.56\textwidth}
\vspace{0pt}
\centering
\begin{minipage}[t][3.6cm][t]{\linewidth}
\centering
\setlength{\tabcolsep}{3pt}
\renewcommand{\arraystretch}{1.35}
\setlength{\arrayrulewidth}{0.8pt}
\small
\begin{tabular}{lrrrrr}
\hline
\textbf{Question Type} & \textbf{\#Q} & \textbf{\#Docs} & \textbf{\shortstack{Corpus Tok.}} & \textbf{\shortstack{{\it AM} Tok.}} & \textbf{\shortstack{\# $C$}} \\
\hline
Single-Sum & 422 & 422 & 9,681,570 & 322,719 & 30 \\
Pair-Comp & 99 & 54 & 1,565,393 &  597,178 & 5 \\
Multi-Comp & 42 & 57 & 1,670,693 & 457,473 & 5 \\
Enumeration & 150 & 64 & 1,728,257 & 427,368 & 7 \\
Temporal & 156 & 1,579 & 1,434,193 & 120,514 & 7 \\
\hline
\textbf{Total} & \textbf{869} & \textbf{2,095} & \textbf{16,080,106} & \textbf{347,963} & \textbf{54} \\
\hline
\end{tabular}
\vfill
\end{minipage}
\captionof{table}{
Data statistics of {\modelname} across question types.
The table uses the abbreviations for the five question types.
$C$ and {\it AM} Tok. denote cluster and average token count of the \textsc{Middle} setting, respectively.
}
\label{tab:adc-data-statistics}
\end{minipage}
\hfill
\begin{minipage}[t]{0.40\textwidth}
\vspace{0pt}
\centering
\begin{minipage}[t][3.6cm][t]{\linewidth}
\centering
\includegraphics[width=\linewidth,trim=8 12 4 0,clip]{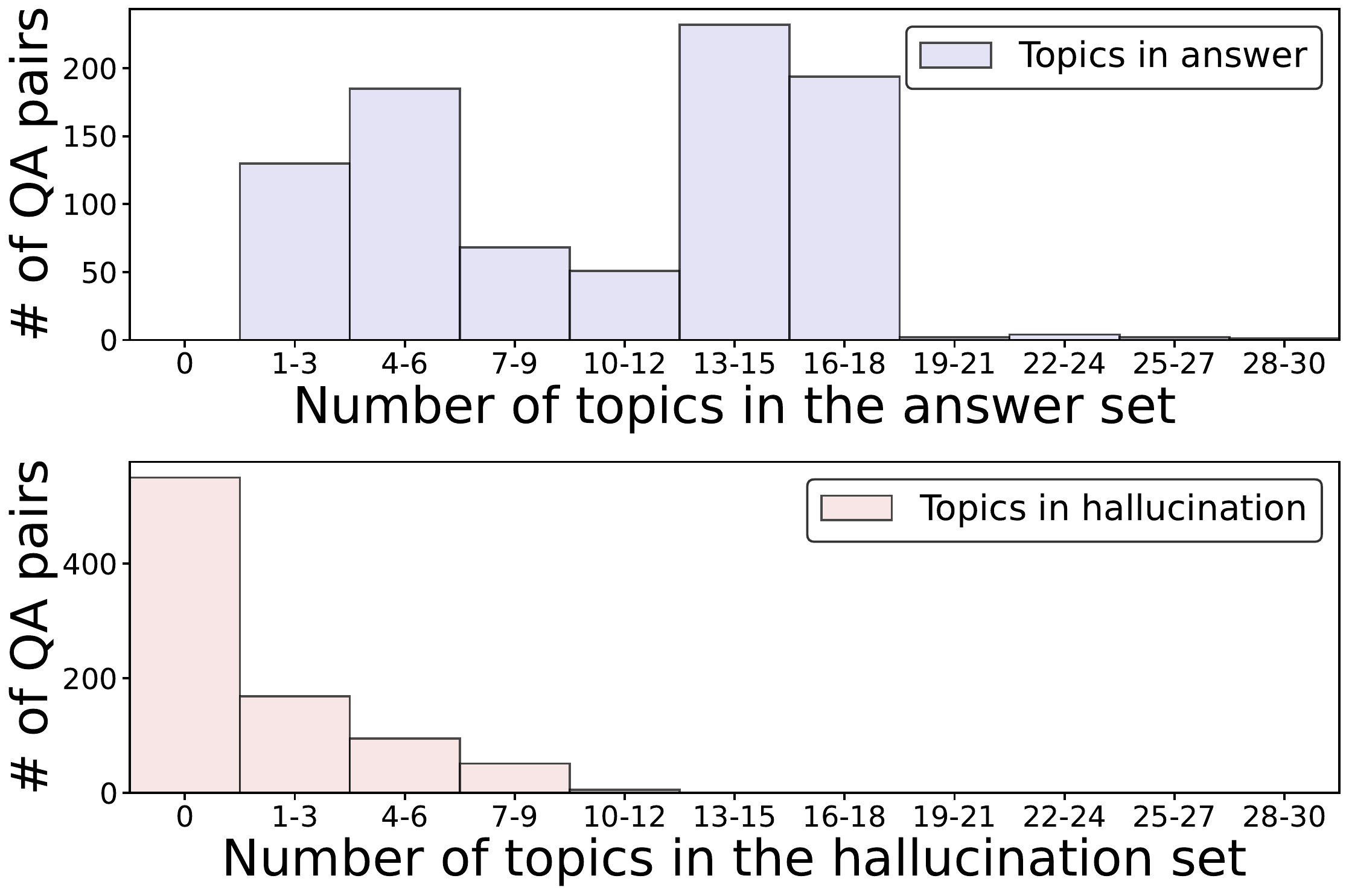}
\end{minipage}
\captionof{figure}{Distribution of topics in the answer and hallucination sets in our {\modelname}.}
\label{fig:distribution}
\end{minipage}
\end{figure}

Table~\ref{tab:adc-data-statistics} summarizes the {\modelname} dataset across the five question types and the retrieval difficulty settings employed in our experiments. 
After filtering and deduplication, {\modelname} comprises 869 QA instances grounded in 2,095 unique source documents and 16,080,106 corpus tokens, organized into 54 clusters under the \textsc{Middle} setting.  
The average token count for this setting is 347,963, as indicated by the \textit{AM} Tok. column, while the largest \textsc{Middle} cluster contains 736,089 tokens. 
\textsc{Single-Sum} constitutes the largest portion of the benchmark, accounting for 48.6\% of the total, while the remaining question types provide complementary coverage of cross-document synthesis and event-centric abstraction.

Another distinctive property of {\modelname} is its topic-set answer representation, as shown in Figure~\ref{fig:distribution}. 
Instead of relying on a single free-form reference, each question is equipped with a set of salient, source-grounded answer topics. Most answers contain 4--18 topics, with none being empty, whereas hallucination sets are concentrated in the 0--3 range, and no hallucination sets exceed 13 topics. 
This design supports joint evaluation of completeness and faithfulness by measuring whether a system covers the major answer topics while avoiding plausible but unsupported ones.

\section{Topic-based Evaluation Method}
\label{sec:reference-based-evaluation}

\subsection{Evaluation Method}
\label{sec:adc-eval-method}

Our core idea is to evaluate {\modelname} answers in the same spirit as grading a composition or a reading-comprehension response by checking whether the answer covers the required key points.
For {\modelname}, a good answer should cover the answer topics in $A$ as completely as possible and avoid hallucinated content.

During the {\modelname} evaluation, we denote the response from a RAG method to a question $Q$ as $Y$.
We then use an LLM-based extractor to convert $Y$ into a response-side topic set $\hat{A}(Q, Y) = \{\hat{\tau}_1, \hat{\tau}_2, \ldots, \hat{\tau}_r\}$.
The extractor is prompted with $Q$, $A$, and $H$ so that it can identify response-side topics with reference to both $A$ and $H$.
If $Y$ contains content that goes beyond both $A$ and $H$, that content is also retained in $\hat{A}(Q, Y)$ rather than discarded during extraction.

We use $\hat{\tau} \equiv \tau$ to denote semantic equivalence between a response-side topic and an answer topic.
We next define the supported response-side topic set and the covered answer topic set as
\[
\small
S(Q, Y) = \{\hat{\tau} \in \hat{A}(Q, Y) \mid \exists \tau \in A,\ \hat{\tau} \equiv \tau\},
\quad
C(Q, Y) = \{\tau \in A \mid \exists \hat{\tau} \in \hat{A}(Q, Y),\ \hat{\tau} \equiv \tau\}.
\]
We then compute topic precision, topic recall, and topic F1 as
\[
\small
\text{T-Prec} = \frac{|S(Q, Y)|}{\max(1, |\hat{A}(Q, Y)|)},
\quad
\text{T-Rec} = \frac{|C(Q, Y)|}{|A|},
\quad
\text{T-F1} = \frac{2\,\text{T-Prec}\cdot\text{T-Rec}}{\text{T-Prec}+\text{T-Rec}}.
\]

To evaluate curated hallucination targets, we define the covered hallucination set as
\[
\small
C_H(Q, Y) = \{h \in H \mid \exists \hat{\tau} \in \hat{A}(Q, Y),\ \hat{\tau} \equiv h\}.
\]
Based on $C_H(Q, Y)$, we define a topic-level hallucination score ($H_{\mathrm{topic}}$) and a response-level hallucination rate ($H_{\mathrm{resp}}$) as
\[
\small
H_{\mathrm{topic}}(Q, Y) = \frac{|C_H(Q, Y)|}{|H|},
\quad
H_{\mathrm{resp}}(Q, Y) = \mathbb{I}\!\left[|C_H(Q, Y)| > 0\right].
\]
Here, $\mathbb{I}[\cdot]$ denotes the indicator function.
A lower $H_{\mathrm{topic}}$ means that the response covers fewer hallucination topics in $H$, while $H_{\mathrm{resp}}(Q, Y)=1$ means that the response contains at least one hallucination topic.
For instances with an empty hallucination set, we set $H_{\mathrm{topic}}(Q,Y)=0$ and $H_{\mathrm{resp}}(Q,Y)=0$.


\subsection{Discussion of the Evaluation Method}
\label{sec:adc-eval-discussion}

We next discuss why this evaluation method fits abstract QA, what it reveals, and how its scope should be interpreted.

\textbf{Fit to abstract RAG.}
{\modelname} answers are structured syntheses rather than isolated facts, so they require evaluation beyond short-form answer matching.
The answer topic set $A$ defines the expected coverage of a good answer, while the hallucination set $H$ provides explicit negative targets for hallucination checking.

\textbf{Scalability and interpretability.}
Our evaluation method avoids exhaustive head-to-head comparisons by scoring each method independently against the same topic references.
This reduces evaluation from $O(n^2)$ pairwise comparisons to $O(n)$ evaluations for $n$ methods, saving both annotation effort and LLM inference cost.
It also produces diagnostic scores because each result can be traced to matched, missing, and hallucination-related topics.
This makes the comparison more interpretable than a single holistic preference label.

\textbf{Scope and limitations.}
Evaluation quality still depends on the quality of the curated topic references and on the reliability of the LLM-based extraction stage.
The hallucination scores should be interpreted as reference-targeted diagnostics rather than exhaustive factuality metrics: the curated hallucination set $H$ captures plausible unsupported topics identified during benchmark construction, but it cannot cover all hallucination patterns that may appear in model responses.
Topic granularity is also a design choice, and topic-set scoring does not fully capture discourse quality, response organization, or stylistic preference.


\section{Experiments}
\label{sec:exp}

In this section, we evaluate representative RAG methods on {\modelname} and show that our benchmark exposes coverage gaps, hallucination risks, and retrieval-scope sensitivity in current RAG systems.

\subsection{Setup}

\textbf{Methods and implementations.}
We evaluate representative RAG methods targeted for abstract QA, covering vanilla, graph-based, and hierarchical retrieval settings.
The compared methods are Vanilla RAG, RAPTOR~\cite{sarthi2024raptor}, HippoRAG~\cite{gutierrez2024hipporag}, GraphRAG~\cite{edge2024local}, LightRAG~\cite{guo2024lightrag}, ArchRAG~\cite{wang2026archrag}, KET-RAG~\cite{huang2025ket}, and HiRAG~\cite{huang2025retrieval}.
For GraphRAG, we evaluate its global and local search, denoted as GGraphRAG and LGraphRAG, respectively.
Similarly, we evaluate the low, high, and hybrid retrieval modes for LightRAG, denoted as LLightRAG, HiLightRAG, and HyLightRAG, respectively.
We use the official implementations for ArchRAG, KET-RAG, and HiRAG, and the Graph-based RAG framework~\cite{zhou2025depth} for the remaining methods.
All methods follow the recommended chunking and retrieval parameters when available.
For Vanilla RAG, we set the chunk size to 512 tokens and retrieve the top 10 chunks.

\textbf{Models and evaluation settings.}
All RAG methods use Qwen3-8B~\cite{yang2025qwen3} as the LLM backbone and nomic-embed-text~\cite{nussbaum2024nomic} as the embedding model.
We use GPT-5.1~\cite{openai2025gpt51systemcard} for evaluation.
All methods are evaluated across three retrieval settings: \textsc{Simple}, \textsc{Middle}, and \textsc{Hard}.
All model outputs are scored with the topic-based evaluation method introduced in Section~\ref{sec:reference-based-evaluation}.
Note that we report only $\text{T-F1}$, $H_{\mathrm{topic}}$, and $H_{\mathrm{resp}}$ in this section, with the remaining metrics and additional details provided in the appendix.

\definecolor{c1}{RGB}{216,174,174} 
\definecolor{c2}{RGB}{224,175,107}  
\definecolor{c3}{RGB}{222,117,123} 
\definecolor{c4}{RGB}{250, 221, 104} 

\definecolor{c5}{RGB}{174,223,172} 
\definecolor{c6}{RGB}{138,170,214}  
\definecolor{c7}{RGB}{248,199,1} 
\definecolor{c8}{RGB}{255,0,127} 

\definecolor{c11}{RGB}{172, 196, 226} 
\definecolor{c12}{RGB}{208, 221, 238} 

\definecolor{c21}{RGB}{163,137,214} 

\definecolor{c9}{RGB}{150,195,125} %
\definecolor{c10}{RGB}{230,189,69} %

\definecolor{c13}{RGB}{115,107,157} %
\definecolor{c14}{RGB}{208,108,157} %

\begin{table*}[ht]
    \centering    
    \caption{Performance comparison of representative RAG methods on {\modelname} under the three retrieval settings and overall, where TF1, HT, and HR denote $\text{T-F1}$, $H_{\mathrm{topic}}$, and $H_{\mathrm{resp}}$, respectively. 
    }
\small
{
\setlength{\tabcolsep}{5pt}
    \begin{tabular}{l|ccc|ccc|ccc|ccc}
\toprule
\multirow{2}{*}{Method} & \multicolumn{3}{c}{Simple} & \multicolumn{3}{c}{Middle} & \multicolumn{3}{c}{Hard} & \multicolumn{3}{c}{Overall} \\
\cmidrule(lr){2-4} \cmidrule(lr){5-7} \cmidrule(lr){8-10} \cmidrule(lr){11-13}
& TF1 & HT & HR & TF1 & HT & HR & TF1 & HT & HR & TF1 & HT & HR \\
\midrule
{Vanilla RAG} & 45.0 & 15.6 & 27.2 & 24.2 & 6.8 & 12.4 & 23.7 & 10.0 & 22.6 & 31.0 & 10.8 & 20.7 \\
{LLightRAG} & 55.0 & 9.8 & 20.7 & 36.7 & 15.9 & 30.4 & 33.9 & 17.9 & 37.0 & 41.8 & 14.5 & 29.4 \\
{HiLightRAG} & 54.0 & 8.2 & 16.7 & 48.4 & 12.4 & 29.9 & 46.0 & 15.4 & 31.3 & 49.5 & 12.0 & 26.0 \\
{HyLightRAG} & 40.7 & 7.6 & 19.5 & 40.0 & 13.8 & 26.7 & 38.4 & 17.5 & 34.9 & 39.7 & 13.0 & 27.0 \\
{LGraphRAG} & 57.8 & 11.0 & 22.3 & 39.3 & 11.9 & 27.5 & 38.3 & 13.9 & 27.9 & 45.1 & 12.3 & 25.9 \\
{GGraphRAG}& 23.1 & 5.2 & 12.7 & 22.5 & 7.2 & 15.5 & 21.8 & 7.7 & 17.4 & 22.5 & 6.7 & 15.2 \\
{HippoRAG} & 61.7 & 17.3 & 35.1 & \cellcolor{best}{56.9} & 27.4 & 45.8 & \cellcolor{best}{51.2} & 17.5 & 35.1 & \cellcolor{best}{56.6} & 20.0 & 38.7 \\
{RAPTOR} & 64.0 & 16.4 & 31.7 & 53.3 & 16.6 & 37.6 & 52.1 & 21.1 & 45.5 & 55.3 & 18.0 & 38.3 \\
{ArchRAG} & 55.2 & 18.9 & 37.5 & 47.6 & 20.7 & 39.4 & 47.0 & 19.3 & 37.6 & 49.9 & 19.6 & 38.2 \\
{KET-RAG} & 32.7 & \cellcolor{best}{4.0} & \cellcolor{best}{6.2} & 27.8 & \cellcolor{best}{3.3} & \cellcolor{best}{6.6} & 12.9 & \cellcolor{best}{5.6} & \cellcolor{best}{12.8} & 24.5 & \cellcolor{best}{4.3} & \cellcolor{best}{8.5} \\
{HiRAG} & \cellcolor{best}{68.9} & 18.1 & 29.2 & 45.2 & 13.8 & 33.0 & 35.7 & 13.4 & 34.0 & 49.9 & 15.1 & 32.1 \\
\bottomrule
    \end{tabular}
}
    \label{tab:retrieval}
\end{table*}

\subsection{Performance of RAG Methods on {\modelname}}
\label{sec:overallExp}

We highlight the best score for each metric in each table in {\setlength{\fboxsep}{0.5pt}\colorbox{best}{red}}, with higher TF1 and lower $H_{\mathrm{topic}}$/$H_{\mathrm{resp}}$ indicating better performance.

\textbf{Overall results.}
Table~\ref{tab:retrieval} reports T-F1, $H_{\mathrm{topic}}$, and $H_{\mathrm{resp}}$ for each method under the three retrieval settings and their overall averages.
In general, most methods degrade as the retrieval scope expands, and stronger topic coverage is often accompanied by higher curated hallucination scores.
HippoRAG achieves the highest overall T-F1, reaching 56.6, but it also has the highest overall $H_{\mathrm{topic}}$ and $H_{\mathrm{resp}}$, at 20.0 and 38.7.
HiRAG obtains a strong overall T-F1 of 49.9 and the best T-F1 under \textsc{Simple} with 68.9, but its overall $H_{\mathrm{resp}}$ remains high at 32.1.
In contrast, KET-RAG yields the lowest overall $H_{\mathrm{topic}}$ and $H_{\mathrm{resp}}$, at 4.3 and 8.5, while its T-F1 drops to 24.5.
These results highlight that {\modelname} separates topic coverage from curated unsupported-topic behavior, rather than treating more verbose or responses with excessive unsupported content as uniformly better.

\textbf{Retrieval-scope comparison.}
Table~\ref{tab:retrieval} also separates results by retrieval scope, allowing direct comparison across \textsc{Simple}, \textsc{Middle}, and \textsc{Hard} for the same method.
LLightRAG drops from 55.0 T-F1 under \textsc{Simple} to 33.9 under \textsc{Hard}, while its $H_{\mathrm{resp}}$ increases from 20.7 to 37.0.
LGraphRAG also declines from 57.8 to 38.3 T-F1, and KET-RAG decreases from 32.7 to 12.9.
Some methods are more stable in coverage, such as HiLightRAG, from 54.0 to 46.0 T-F1, and HippoRAG, from 61.7 to 51.2 T-F1.
However, broader retrieval scopes still increase hallucination risk for several methods, as shown by RAPTOR's $H_{\mathrm{resp}}$ rising from 31.7 to 45.5.
These changes show that {\modelname}'s retrieval settings not only increase corpus size, but also test whether a method can keep the answer within the designated evidence scope.

\begin{table*}[t]
    \setlength{\belowcaptionskip}{-0.2cm}
    \centering    
    \caption{Performance comparison of representative RAG methods on {\modelname} across the five question types, where TF1, HT, and HR denote $\text{T-F1}$, $H_{\mathrm{topic}}$, and $H_{\mathrm{resp}}$, respectively.}
\small
{
\renewcommand{\arraystretch}{0.99}
\setlength{\tabcolsep}{3pt}
    \begin{tabular}{l|ccc|ccc|ccc|ccc|ccc}
\toprule 
\multirow{2}{*}{Method} & \multicolumn{3}{c}{Single-Sum} & \multicolumn{3}{c}{Pair-Comp} & \multicolumn{3}{c}{Multi-Comp} & \multicolumn{3}{c}{Enumeration} & \multicolumn{3}{c}{Temporal} \\
\cmidrule(lr){2-4} \cmidrule(lr){5-7} \cmidrule(lr){8-10} \cmidrule(lr){11-13} \cmidrule(lr){14-16}
& TF1 & HT & HR & TF1 & HT & HR & TF1 & HT & HR & TF1 & HT & HR & TF1 & HT & HR \\
\midrule
{Vanilla RAG} & 14.8 & 1.3  & 1.3 & 63.7 & 16.4 & 19.1 & 52.3 & 13.0 & 13.9 & 56.3 & 16.7 & 16.7 & 23.8 & 12.2 & 33.3 \\
{LLightRAG} & 48.1 & 3.8 & 3.8 & 52.3 & 16.5 & 21.3 & 48.4 & 10.7 & 15.3 & 41.6 & 16.7 & 16.7 & 16.6 & 20.4 & 52.2 \\
{HiLightRAG} & 56.9 & 4.6 & 5.5 & 59.1 & 12.3 & 18.4 & 56.9 & 13.5 & 23.1 & 56.6 & 19.4 & 19.4 & 14.5 & 14.4 & 39.6 \\
{HyLightRAG} & 44.5 & 2.5 & 3.3 & 43.9 & 12.2 & 17.8 & 44.4 & 6.9 & 9.3 & 42.4 & \cellcolor{best}{0.0} & \cellcolor{best}{0.0} & 20.5 & 17.4 & 41.2 \\
{LGraphRAG} & 52.7 & 2.5 & 4.3 & 48.7 & 12.4 & 16.3 & 59.1 & 13.0 & 20.8 & 48.3 & \cellcolor{best}{0.0} & \cellcolor{best}{0.0} & 15.8 & 17.2 & 43.1 \\
{GGraphRAG} & 35.2 & 0.6 & 2.0 & 12.9 & \cellcolor{best}{2.3} & \cellcolor{best}{5.7} & 11.2 & \cellcolor{best}{0.3} & \cellcolor{best}{1.9} & 15.2 & \cellcolor{best}{0.0} & \cellcolor{best}{0.0} & 7.2 & 13.6 & 30.5 \\
{HippoRAG} & \cellcolor{best}{66.9} & 4.4 & 4.7 & 66.1 & 25.8 & 31.6 & \cellcolor{best}{68.6} & 19.4 & 29.2 & 61.2 & 8.3 & 8.3 & 15.1 & 27.6 & 63.2 \\
{RAPTOR} & 65.2 & 6.4 & 6.4 & \cellcolor{best}{66.9} & 15.1 & 25.4 & 68.1 & 13.6 & 21.9 & 58.7 & 16.7 & 16.7 & 14.3 & 22.0 & 52.7 \\
{ArchRAG} & 58.9 & 4.5 & 6.5 & 60.6 & 21.9 & 27.6 & 62.1 & 16.7 & 27.8 & 49.6 & 33.3 & 33.3 & 15.9 & 26.5 & 62.1 \\
{KET-RAG} & 29.3 & \cellcolor{best}{0.5} & \cellcolor{best}{1.0} & 25.5 & 5.3 & 8.5 & 14.1 & 2.1 & 4.2 & 34.3 & 16.7 & 16.7 & 3.9 & \cellcolor{best}{5.2} & \cellcolor{best}{12.0} \\
{HiRAG} & 49.3 & 5.8 & 5.8 & 54.8 & 12.0 & 14.8 & 43.5 & 7.6 & 10.4 & \cellcolor{best}{68.1} & \cellcolor{best}{0.0} & \cellcolor{best}{0.0} & \cellcolor{best}{33.5} & 22.1 & 56.0 \\
\bottomrule
    \end{tabular}
}
    \label{tab:tasks}
\end{table*}

\textbf{Question-type comparison.}
Table~\ref{tab:tasks} reports the performance of different RAG methods across the five {\modelname} question types.
Single-Sum favors methods that summarize a focused source, with HippoRAG and RAPTOR reaching 66.9 and 65.2 T-F1.
On \textsc{Enumeration}, HiRAG achieves the highest T-F1 of 68.1 and several methods obtain 0 $H_{\mathrm{topic}}$ and $H_{\mathrm{resp}}$, because only 4 Enumeration QA instances contain hallucination sets.
\textsc{Temporal} is the most difficult question type, with the best T-F1 reaching 33.5 and $H_{\mathrm{resp}}$ remaining high, peaking at 63.2 for HippoRAG.
This difficulty mainly comes from the much larger document sets in Temporal and the lack of temporal evidence selection mechanisms in most compared methods.
These variations across question types show that {\it {\modelname} tests distinct forms of abstract QA over documents rather than only scaling up retrieval difficulty}.

\subsection{Comparison with Head-to-Head Evaluation}

To compare topic-based evaluation with the commonly used head-to-head evaluation, we randomly sample 200 questions under the \textsc{Simple} setting and evaluate both ranking consistency and evaluation cost.
In the head-to-head setting, each cell represents the win rate of the row method against the column method; \textit{single-order} evaluates each ordered pair once, while \textit{bidirectional} also evaluates the reversed order and reports the average.
We report both directions because head-to-head evaluation is sensitive to the presentation order of RAG model responses, and the resulting win rates can vary noticeably when the method order is reversed.
The detailed bidirectional results are provided in the appendix, which further suggests that head-to-head evaluation may yield unstable conclusions even under the same question set.
As shown in Figure~\ref{fig:abstract-winrate} and Table~\ref{tab:main-setting}, {\modelname} produces rankings that are consistent with head-to-head results, with RAPTOR performing best and KET-RAG performing worst.
Figure~\ref{fig:eval-cost} further shows that {\modelname} reduces evaluation time from $154,784$s to $7,473$s and token cost by over $21\times$, because each response is scored independently against fixed topic sets rather than compared against every other method.
Note that we report only the overall results for five selected methods, and the complete results are provided in the appendix.


\pgfplotstableread[row sep=\\,col sep=&]{
metric & H2HS & H2HD & ADC & H2HSLabel & H2HDLabel & ADCLabel \\
2 & 0.491 & 1.000 & 0.048 & 76.0K & 154.8K & 7.5K \\
4 & 0.4999 & 1.000 & 0.046 & 18.3M & 36.6M & 1.7M \\
}\evalcostdata

\begin{figure*}[ht]
    \setlength{\belowcaptionskip}{-0.2cm}
\centering
\begin{minipage}[t]{0.32\textwidth}
\vspace{0pt}
\centering
\begin{minipage}[t][2.5cm][t]{\linewidth}
\centering
    \resizebox{0.95\textwidth}{!}{ 
\huge
\renewcommand{\arraystretch}{1.12}
    \begin{tabular}{c*{5}{c}}
  & RA & HR & GL & HY & KR \\ 
{RA} & \colorcell{50} & \colorcell{53} & \colorcell{74} & \colorcell{87} & \colorcell{99} \\ 
{HR} & \colorcell{47} & \colorcell{50} & \colorcell{66} & \colorcell{76} & \colorcell{98} \\ 
{GL} & \colorcell{26} & \colorcell{34} & \colorcell{50} & \colorcell{69} & \colorcell{95} \\ 
{HY} & \colorcell{13} & \colorcell{24} & \colorcell{31} & \colorcell{50} & \colorcell{88} \\ 
{KR} & \colorcell{1} & \colorcell{2} & \colorcell{5} & \colorcell{12} & \colorcell{50} \\ 
    \end{tabular}
    }
\vfill
\end{minipage}
\captionof{figure}{Head-to-head (bidirectional, overall) win-rate comparison, using the same method abbreviations as in Table~\ref{tab:main-setting}.}
\label{fig:abstract-winrate}
\end{minipage}
\hfill
\begin{minipage}[t]{0.32\textwidth}
\vspace{0pt}
\centering
\begin{minipage}[t][2.5cm][t]{\linewidth}
\centering
{
\small
\setlength{\tabcolsep}{4pt}
\renewcommand{\arraystretch}{1.2}
\begin{tabular}{lccc}
\hline
Method & TF1 & HT & HR  \\
\hline
RAPTOR      & 62.8 & 9.4 & 20.0 \\
HippoRAG    & 61.7 & 15.4 & 35.5 \\
LGraphRAG   & 55.6 & 7.5 & 16.3\\
HyLightRAG  & 40.8 & 6.4 & 13.8\\
KET-RAG     & 37.7 & 1.1  & 2.1\\
\hline
\end{tabular}
}
\vfill
\end{minipage}
\captionof{table}{{\modelname} evaluation on 200 QA instances. TF1, HT, and HR denote $\text{T-F1}$, $H_{\mathrm{topic}}$, and $H_{\mathrm{resp}}$, respectively.}
\label{tab:main-setting}
\end{minipage}
\hfill
\begin{minipage}[t]{0.34\textwidth}
\vspace{0pt}
\centering
\begin{minipage}[t][2.5cm][t]{\linewidth}
\centering
\begin{tikzpicture}[scale=0.97]
\begin{axis}[
    ybar,
    bar width=0.44cm,
    width=1.15\linewidth,
    height=0.66\linewidth,
    ymin=0, ymax=1.3,
    xmin=1.6, xmax=4.4,
    xtick=data,
    xticklabels={Time, Tokens},
    xticklabel style={font=\footnotesize},
    ytick={0,0.5,1.0},
    yticklabel style={font=\footnotesize},
    ylabel={\textbf{\footnotesize Relative Cost}},
    ylabel style={yshift=-16pt},
    ymajorgrids=true,
    grid style=dashed,
    tick align=inside,
    enlarge x limits=0.35,
    legend style={
        at={(0.5,1.02)},
        anchor=south,
        legend columns=3,
        draw=none,
        font=\footnotesize
    },
    legend image code/.code={
        \draw [#1, line width=0.5pt] (0cm,-0.1cm) rectangle (0.25cm,0.18cm);
    },
    extra y ticks = {1.0},
    extra y tick style={grid=major, grid style={line width=1.5pt, black, dashed}},
    every axis plot/.append style={line width = 1.0pt},
    every axis/.append style={line width = 1.5pt},
    nodes near coords style={
        font=\tiny,      
        color=black,
        anchor=south,    
        yshift=0pt,     
    },
]

\addplot[
    fill=c2,
    postaction={pattern=vertical lines, pattern color=black!80},
    draw=black,
    nodes near coords=\pgfplotspointmeta, 
    point meta=explicit symbolic
] table[x=metric, y=H2HD, meta=H2HDLabel] {\evalcostdata};

\addplot[
    fill=c6,
    postaction={pattern=north west lines, pattern color=black!80},
    draw=black,
    nodes near coords=\pgfplotspointmeta, 
    point meta=explicit symbolic
] table[x=metric, y=H2HS, meta=H2HSLabel] {\evalcostdata};

\addplot[
    fill=c3,
    draw=black,
    nodes near coords=\pgfplotspointmeta, 
    point meta=explicit symbolic
] table[x=metric, y=ADC, meta=ADCLabel] {\evalcostdata};

\legend{H2H-D, H2H-S, {\modelname}}
\end{axis}
\end{tikzpicture}
\vspace{-0.2cm}

\end{minipage}

\captionof{figure}{Normalized time and token costs relative to Head-to-Head (bidirectional). Absolute values are annotated above the bars.}
\label{fig:eval-cost}
\end{minipage}

\end{figure*}

\subsection{Dataset Analysis}

Benchmark diversity is particularly important for abstract QA, as it exposes systems to a broader range of linguistic variation and synthesis demands, and therefore provides a more realistic test of their generalization ability~\cite{carmel2025liverag,han2014big}.
Table~\ref{tab:dataset-analysis} compares {\modelname} with representative RAG datasets~\cite{rabinovich2023predicting,tang2024multihop,han2024rag,qian2025memorag,carmel2025liverag} using three diversity metrics from prior text generation studies~\cite{shaib2024standardizing}: Q\&A n-gram diversity (Q\&A-NGD), answer cluster entropy (A-CE), and answer embedding homogeneity score (A-HS), together with average question and answer lengths; further details are provided in the appendix.
Answer-side metrics are unavailable for UltraDomain because it provides only questions without answers, while A-HS is not applicable to MultiHop-RAG because each question has only a single target answer.
Although {\modelname} does not achieve the highest Q\&A-NGD, its score of 2.86 remains close to LiveRAG's 2.97, indicating that its questions are still lexically diverse rather than generated from narrow templates.
{\modelname} also attains the best A-CE and A-HS, because its topic-set answers are designed to cover more salient aspects of the source documents and therefore span more diverse answer-side semantic regions.
Overall, these results show that {\modelname} provides strong diversity on both the question and answer sides, with particularly rich semantic coverage and a balanced semantic distribution across its topic-set answers.

\begin{table*}[ht]
\centering
\caption{Comparison of representative RAG datasets and {\modelname}.
$\uparrow$/$\downarrow$ mark higher/lower-is-better, and the best and second-best results are marked in {\bf bold} and \underline{underlined}, respectively.
}
\label{tab:dataset-analysis}
\small
\setlength{\tabcolsep}{6pt}
\begin{tabular}{lccccc}
\toprule
Dataset & {Q\&A-NGD $\uparrow$} & { Q-Length} & { A-CE $\uparrow$} & { A-HS $\downarrow$} & {A-Length} \\
\midrule
PopQA~\cite{rabinovich2023predicting} & 2.37 & 6.70 & 3.76& 0.45 & 3.68 \\
MultiHop-RAG~\cite{tang2024multihop} & 1.99 & 46.51 & 2.35& / & 1.31 \\
RAG-QA Arena~\cite{han2024rag}      & 2.82 & 9.89 & 3.83& 0.37 & 80.16 \\
UltraDomain~\cite{qian2025memorag}  & 2.31 & 12.56 &  /& / & / \\
LiveRAG~\cite{carmel2025liverag}    & \textbf{2.97} & 15.09 & \underline{3.84} & \underline{0.36} & 60.61 \\
\midrule
{\bf {\modelname} (Ours)}                    & \underline{2.86} & 16.25 & \textbf{3.85} & \textbf{0.14} & 17.17 \\
\bottomrule
\end{tabular}
\end{table*}

\section{Conclusion}
\label{sec:conclusion}

In this paper, we introduce {\modelname}, a benchmark for abstract question answering over documents.
{\modelname} provides curated QA instances with answer topic sets, hallucination sets, and grounded evidence, enabling evaluation beyond short-answer matching.
We further present a topic-based evaluation method that measures topic coverage and curated unsupported-topic matches in a more interpretable and more scalable way than head-to-head judging.
Experiments on representative RAG baselines show that they still struggle on {\modelname}, especially when the retrieval scope expands from focused evidence to broader and more distracting corpora.
These results position {\modelname} as a practical benchmark and evaluation testbed for studying more capable RAG methods with realistic retrieval conditions for solving the abstract QA task.



\newpage

\addcontentsline{toc}{section}{Bibliography}
\bibliographystyle{unsrtnat}
{\bibliography{references}}

@article{sarthi2024raptor,
  title={Raptor: Recursive abstractive processing for tree-organized retrieval},
  author={Sarthi, Parth and Abdullah, Salman and Tuli, Aditi and Khanna, Shubh and Goldie, Anna and Manning, Christopher D},
  journal={arXiv preprint arXiv:2401.18059},
  year={2024}
}

@article{yang2018hotpotqa,
  title={HotpotQA: A dataset for diverse, explainable multi-hop question answering},
  author={Yang, Zhilin and Qi, Peng and Zhang, Saizheng and Bengio, Yoshua and Cohen, William W and Salakhutdinov, Ruslan and Manning, Christopher D},
  journal={arXiv preprint arXiv:1809.09600},
  year={2018}
}

@article{zheng2023judging,
  title={Judging llm-as-a-judge with mt-bench and chatbot arena},
  author={Zheng, Lianmin and Chiang, Wei-Lin and Sheng, Ying and Zhuang, Siyuan and Wu, Zhanghao and Zhuang, Yonghao and Lin, Zi and Li, Zhuohan and Li, Dacheng and Xing, Eric and others},
  journal={Advances in Neural Information Processing Systems},
  volume={36},
  pages={46595--46623},
  year={2023}
}

@article{tang2024multihop,
  title={Multihop-rag: Benchmarking retrieval-augmented generation for multi-hop queries},
  author={Tang, Yixuan and Yang, Yi},
  journal={arXiv preprint arXiv:2401.15391},
  year={2024}
}

@article{lewis2020retrieval,
  title={Retrieval-augmented generation for knowledge-intensive nlp tasks},
  author={Lewis, Patrick and Perez, Ethan and Piktus, Aleksandra and Petroni, Fabio and Karpukhin, Vladimir and Goyal, Naman and K{\"u}ttler, Heinrich and Lewis, Mike and Yih, Wen-tau and Rockt{\"a}schel, Tim and others},
  journal={Advances in Neural Information Processing Systems},
  volume={33},
  pages={9459--9474},
  year={2020}
}

@article{edge2024local,
  title={From local to global: A graph rag approach to query-focused summarization},
  author={Edge, Darren and Trinh, Ha and Cheng, Newman and Bradley, Joshua and Chao, Alex and Mody, Apurva and Truitt, Steven and Larson, Jonathan},
  journal={arXiv preprint arXiv:2404.16130},
  year={2024}
}

@article{gutierrez2024hipporag,
  title={HippoRAG: Neurobiologically Inspired Long-Term Memory for Large Language Models},
  author={Guti{\'e}rrez, Bernal Jim{\'e}nez and Shu, Yiheng and Gu, Yu and Yasunaga, Michihiro and Su, Yu},
  journal={arXiv preprint arXiv:2405.14831},
  year={2024}
}

@misc{nussbaum2024nomic,
      title={Nomic Embed: Training a Reproducible Long Context Text Embedder}, 
      author={Zach Nussbaum and John X. Morris and Brandon Duderstadt and Andriy Mulyar},
      year={2024},
      eprint={2402.01613},
      archivePrefix={arXiv},
      primaryClass={cs.CL}
}

@article{guo2024lightrag,
  title={LightRAG: Simple and Fast Retrieval-Augmented Generation},
  author={Guo, Zirui and Xia, Lianghao and Yu, Yanhua and Ao, Tu and Huang, Chao},
  journal={arXiv e-prints},
  pages={arXiv--2410},
  year={2024}
}

@article{dasigi2021dataset,
  title={A dataset of information-seeking questions and answers anchored in research papers},
  author={Dasigi, Pradeep and Lo, Kyle and Beltagy, Iz and Cohan, Arman and Smith, Noah A and Gardner, Matt},
  journal={arXiv preprint arXiv:2105.03011},
  year={2021}
}

@article{wang2024mineru,
  title={Mineru: An open-source solution for precise document content extraction},
  author={Wang, Bin and Xu, Chao and Zhao, Xiaomeng and Ouyang, Linke and Wu, Fan and Zhao, Zhiyuan and Xu, Rui and Liu, Kaiwen and Qu, Yuan and Shang, Fukai and others},
  journal={arXiv preprint arXiv:2409.18839},
  year={2024}
}

@article{yang2025qwen3,
  title={Qwen3 technical report},
  author={Yang, An and Li, Anfeng and Yang, Baosong and Zhang, Beichen and Hui, Binyuan and Zheng, Bo and Yu, Bowen and Gao, Chang and Huang, Chengen and Lv, Chenxu and others},
  journal={arXiv preprint arXiv:2505.09388},
  year={2025}
}

@article{zhou2025depth,
  title={In-depth Analysis of Graph-based RAG in a Unified Framework},
  author={Zhou, Yingli and Su, Yaodong and Sun, Youran and Wang, Shu and Wang, Taotao and He, Runyuan and Zhang, Yongwei and Liang, Sicong and Liu, Xilin and Ma, Yuchi and others},
  journal={arXiv preprint arXiv:2503.04338},
  year={2025}
}

@inproceedings{wang2026archrag,
  title={Archrag: Attributed community-based hierarchical retrieval-augmented generation},
  author={Wang, Shu and Fang, Yixiang and Zhou, Yingli and Liu, Xilin and Ma, Yuchi},
  booktitle={Proceedings of the AAAI Conference on Artificial Intelligence},
  volume={40},
  number={19},
  pages={15868--15876},
  year={2026}
}

@inproceedings{huang2025ket,
  title={Ket-rag: A cost-efficient multi-granular indexing framework for graph-rag},
  author={Huang, Yiqian and Zhang, Shiqi and Xiao, Xiaokui},
  booktitle={Proceedings of the 31st ACM SIGKDD Conference on Knowledge Discovery and Data Mining V. 2},
  pages={1003--1012},
  year={2025}
}

@article{epstein2020mapping,
  title={Mapping and taking stock of the personal informatics literature},
  author={Epstein, Daniel A and Caldeira, Clara and Figueiredo, Mayara Costa and Lu, Xi and Silva, Lucas M and Williams, Lucretia and Lee, Jong Ho and Li, Qingyang and Ahuja, Simran and Chen, Qiuer and others},
  journal={Proceedings of the ACM on Interactive, Mobile, Wearable and Ubiquitous Technologies},
  volume={4},
  number={4},
  pages={1--38},
  year={2020},
  publisher={ACM New York, NY, USA}
}

@inproceedings{soergel2013openreview,
  title={Open Scholarship and Peer Review: a Time for Experimentation},
  author={Soergel, David and Saunders, Adam and McCallum, Andrew},
  booktitle={ICML 2013 Workshop on Peer Reviewing and Publishing Models},
  year={2013},
  url={https://openreview.net/forum?id=xf0zSBd2iufMg}
}

@article{huang2025retrieval,
  title={Retrieval-augmented generation with hierarchical knowledge},
  author={Huang, Haoyu and Huang, Yongfeng and Yang, Junjie and Pan, Zhenyu and Chen, Yongqiang and Ma, Kaili and Chen, Hongzhi and Cheng, James},
  journal={arXiv preprint arXiv:2503.10150},
  year={2025}
}

@article{hurst2024gpt,
  title={Gpt-4o system card},
  author={Hurst, Aaron and Lerer, Adam and Goucher, Adam P and Perelman, Adam and Ramesh, Aditya and Clark, Aidan and Ostrow, AJ and Welihinda, Akila and Hayes, Alan and Radford, Alec and others},
  journal={arXiv preprint arXiv:2410.21276},
  year={2024}
}

@misc{openai2025gpt51systemcard,
  title        = {GPT-5.1 Instant and GPT-5.1 Thinking System Card Addendum},
  author       = {{OpenAI}},
  year         = {2025},
  howpublished = {\url{https://openai.com/index/gpt-5-system-card-addendum-gpt-5-1/}},
  note         = {Accessed: 2026-04-14}
}

@inproceedings{bai2024longbench,
  title={Longbench: A bilingual, multitask benchmark for long context understanding},
  author={Bai, Yushi and Lv, Xin and Zhang, Jiajie and Lyu, Hongchang and Tang, Jiankai and Huang, Zhidian and Du, Zhengxiao and Liu, Xiao and Zeng, Aohan and Hou, Lei and others},
  booktitle={Proceedings of the 62nd annual meeting of the association for computational linguistics (volume 1: Long papers)},
  pages={3119--3137},
  year={2024}
}

@inproceedings{baumgartner2025peerqa,
  title={Peerqa: A scientific question answering dataset from peer reviews},
  author={Baumg{\"a}rtner, Tim and Briscoe, Ted and Gurevych, Iryna},
  booktitle={Proceedings of the 2025 Conference of the Nations of the Americas Chapter of the Association for Computational Linguistics: Human Language Technologies (Volume 1: Long Papers)},
  pages={508--544},
  year={2025}
}

@inproceedings{stelmakh2022asqa,
  title={ASQA: Factoid questions meet long-form answers},
  author={Stelmakh, Ivan and Luan, Yi and Dhingra, Bhuwan and Chang, Ming-Wei},
  booktitle={Proceedings of the 2022 Conference on Empirical Methods in Natural Language Processing},
  pages={8273--8288},
  year={2022}
}

@inproceedings{fan2019eli5,
  title={ELI5: Long form question answering},
  author={Fan, Angela and Jernite, Yacine and Perez, Ethan and Grangier, David and Weston, Jason and Auli, Michael},
  booktitle={Proceedings of the 57th annual meeting of the association for computational linguistics},
  pages={3558--3567},
  year={2019}
}

@inproceedings{izacard2021leveraging,
  title={Leveraging passage retrieval with generative models for open domain question answering},
  author={Izacard, Gautier and Grave, Edouard},
  booktitle={Proceedings of the 16th conference of the european chapter of the association for computational linguistics: main volume},
  pages={874--880},
  year={2021}
}

@inproceedings{liu2023g,
  title={G-eval: NLG evaluation using gpt-4 with better human alignment},
  author={Liu, Yang and Iter, Dan and Xu, Yichong and Wang, Shuohang and Xu, Ruochen and Zhu, Chenguang},
  booktitle={Proceedings of the 2023 conference on empirical methods in natural language processing},
  pages={2511--2522},
  year={2023}
}

@inproceedings{chen2026pathrag,
  title={Pathrag: Pruning graph-based retrieval augmented generation with relational paths},
  author={Chen, Boyu and Guo, Zirui and Yang, Zidan and Chen, Yuluo and Chen, Junze and Liu, Zhenghao and Shi, Chuan and Yang, Cheng},
  booktitle={Proceedings of the AAAI Conference on Artificial Intelligence},
  volume={40},
  number={36},
  pages={30183--30191},
  year={2026}
}

@inproceedings{cai2024forag,
  title={Forag: Factuality-optimized retrieval augmented generation for web-enhanced long-form question answering},
  author={Cai, Tianchi and Tan, Zhiwen and Song, Xierui and Sun, Tao and Jiang, Jiyan and Xu, Yunqi and Zhang, Yinger and Gu, Jinjie},
  booktitle={Proceedings of the 30th ACM SIGKDD Conference on Knowledge Discovery and Data Mining},
  pages={199--210},
  year={2024}
}

@inproceedings{min2023factscore,
  title={Factscore: Fine-grained atomic evaluation of factual precision in long form text generation},
  author={Min, Sewon and Krishna, Kalpesh and Lyu, Xinxi and Lewis, Mike and Yih, Wen-tau and Koh, Pang and Iyyer, Mohit and Zettlemoyer, Luke and Hajishirzi, Hannaneh},
  booktitle={Proceedings of the 2023 Conference on Empirical Methods in Natural Language Processing},
  pages={12076--12100},
  year={2023}
}

@inproceedings{fabbri2022qafacteval,
  title={QAFactEval: Improved QA-based factual consistency evaluation for summarization},
  author={Fabbri, Alexander Richard and Wu, Chien-Sheng and Liu, Wenhao and Xiong, Caiming},
  booktitle={Proceedings of the 2022 Conference of the North American Chapter of the Association for Computational Linguistics: Human Language Technologies},
  pages={2587--2601},
  year={2022}
}

@inproceedings{scialom2021questeval,
  title={QuestEval: Summarization asks for fact-based evaluation},
  author={Scialom, Thomas and Dray, Paul-Alexis and Lamprier, Sylvain and Piwowarski, Benjamin and Staiano, Jacopo and Wang, Alex and Gallinari, Patrick},
  booktitle={Proceedings of the 2021 conference on empirical methods in natural language processing},
  pages={6594--6604},
  year={2021}
}

@inproceedings{durmus2020feqa,
  title={FEQA: A question answering evaluation framework for faithfulness assessment in abstractive summarization},
  author={Durmus, Esin and He, He and Diab, Mona},
  booktitle={Proceedings of the 58th annual meeting of the association for computational linguistics},
  pages={5055--5070},
  year={2020}
}

@article{zhang2019bertscore,
  title={Bertscore: Evaluating text generation with bert},
  author={Zhang, Tianyi and Kishore, Varsha and Wu, Felix and Weinberger, Kilian Q and Artzi, Yoav},
  journal={arXiv preprint arXiv:1904.09675},
  year={2019}
}

@inproceedings{lin2004rouge,
  title={Rouge: A package for automatic evaluation of summaries},
  author={Lin, Chin-Yew},
  booktitle={Text summarization branches out},
  pages={74--81},
  year={2004}
}

@inproceedings{han2024rag,
  title={Rag-qa arena: Evaluating domain robustness for long-form retrieval augmented question answering},
  author={Han, Rujun and Zhang, Yuhao and Qi, Peng and Xu, Yumo and Wang, Jenyuan and Liu, Lan and Wang, William Yang and Min, Bonan and Castelli, Vittorio},
  booktitle={Proceedings of the 2024 conference on empirical methods in natural language processing},
  pages={4354--4374},
  year={2024}
}

@article{yang2024crag,
  title={Crag-comprehensive rag benchmark},
  author={Yang, Xiao and Sun, Kai and Xin, Hao and Sun, Yushi and Bhalla, Nikita and Chen, Xiangsen and Choudhary, Sajal and Gui, Rongze D and Jiang, Ziran W and Jiang, Ziyu and others},
  journal={Advances in Neural Information Processing Systems},
  volume={37},
  pages={10470--10490},
  year={2024}
}

@article{carmel2025liverag,
  title={LiveRAG: A diverse Q\&A dataset with varying difficulty level for RAG evaluation},
  author={Carmel, David and Filice, Simone and Horowitz, Guy and Maarek, Yoelle and Shtoff, Alex and Somekh, Oren and Tavory, Ran},
  journal={arXiv preprint arXiv:2511.14531},
  year={2025}
}

@inproceedings{rabinovich2023predicting,
  title={Predicting question-answering performance of large language models through semantic consistency},
  author={Rabinovich, Ella and Ackerman, Samuel and Raz, Orna and Farchi, Eitan and Tavor, Ateret Anaby},
  booktitle={Proceedings of the Third Workshop on Natural Language Generation, Evaluation, and Metrics (GEM)},
  pages={138--154},
  year={2023}
}

@inproceedings{qian2025memorag,
  title={Memorag: Boosting long context processing with global memory-enhanced retrieval augmentation},
  author={Qian, Hongjin and Liu, Zheng and Zhang, Peitian and Mao, Kelong and Lian, Defu and Dou, Zhicheng and Huang, Tiejun},
  booktitle={Proceedings of the ACM on Web Conference 2025},
  pages={2366--2377},
  year={2025}
}

@inproceedings{han2014big,
  title={On big data benchmarking},
  author={Han, Rui and Lu, Xiaoyi and Xu, Jiangtao},
  booktitle={Workshop on Big Data Benchmarks, Performance Optimization, and Emerging Hardware},
  pages={3--18},
  year={2014},
  organization={Springer}
}

@article{shaib2024standardizing,
  title={Standardizing the measurement of text diversity: A tool and a comparative analysis of scores},
  author={Shaib, Chantal and Govindarajan, Venkata S and Barrow, Joe and Sun, Jiuding and Siu, Alexa F and Wallace, Byron C and Nenkova, Ani},
  journal={arXiv preprint arXiv:2403.00553},
  year={2024}
}

@inproceedings{kwon2023efficient,
  title={Efficient Memory Management for Large Language Model Serving with PagedAttention},
  author={Woosuk Kwon and Zhuohan Li and Siyuan Zhuang and Ying Sheng and Lianmin Zheng and Cody Hao Yu and Joseph E. Gonzalez and Hao Zhang and Ion Stoica},
  booktitle={Proceedings of the ACM SIGOPS 29th Symposium on Operating Systems Principles},
  year={2023}
}

@misc{ollama2024,
  title        = {Ollama},
  author       = {{Ollama}},
  year         = {2024},
  howpublished = {\url{https://github.com/ollama/ollama}},
  note         = {Accessed: 2026-05-03}
}

@String{Springer = "Springer-Verlag" }

\newpage


\appendix

\clearpage

\section{Additional Details of {\modelname}}
\label{sec:app-data}

\subsection{Examples of {\modelname} benchmark}

\begin{table}[h]
    \centering
\caption{Example QA instance of {\modelname} benchmark across the five question types.}
    \begin{tabular}{lp{5.5cm}p{5cm}}
\toprule
{\bf Question Type} &  {\bf Question} & {\bf Answer Topic Set} \\
\midrule
Single-Sum & Please summarize the paper Leveraging Large Language Models for Multiple Choice Question Answering. & [Multiple Choice Question Answering (MCQA), Cloze Prompting (CP), $\cdots$] (16 topics) \\

Pair-Comp & How do NMT-based and LLM-based approaches differ in their methodologies and capabilities for program repair? &  [Translation Task, Sequence-to-Sequence, Copy Mechanism, Limited PL Knowledge, Attention Mechanism, $\cdots$] (17 topics)\\

Multi-Comp & How do single-functional, dual-functional, and triple-functional signal designs for ISCC compare in terms of their approach to signal integration and system complexity? & [Independent Signals, Severe Interference, High Complexity Beamforming, Sensing and AirComp Signals, $\cdots$] (10 topics)\\

Enumeration & What mechanisms do gradient-based jailbreak attacks employ to manipulate LLMs, and what are the potential security consequences of these mechanisms?  & [GCG, AutoDAN, ARCA, Append adversarial suffix/prefix to prompts, $\cdots$] (7 topics)\\

Temporal & What are the implications of US-China economic interactions on consumer costs and national economic resilience? & [Increased costs for consumers; Inflation, Rising prices affecting households, $\cdots$] (9 topics)\\
\bottomrule
    \end{tabular}
    \label{tab:adc_example}
\end{table}

\subsection{Discussion of {\modelname}}

{\bf Broader impacts.} 
{\modelname} is intended to support more transparent and reliable evaluation of document-grounded QA and RAG systems, especially for abstract questions that require synthesis across long or multiple documents. 
By providing reference topic sets, hallucination sets, and evidence metadata, the benchmark may help researchers diagnose missing coverage and unsupported content, reduce reliance on opaque pairwise preferences, and improve the reproducibility of RAG evaluation. 
At the same time, benchmark results should not be interpreted as a complete measure of real-world system safety or usefulness.
Systems that perform well on {\modelname} may still propagate biases or inaccuracies from source documents, fail under deployment-specific constraints, or produce unsupported content outside the curated hallucination sets. 
In high-stakes domains such as scientific, legal, medical, or policy decision support, outputs from RAG systems should therefore remain subject to domain-expert review and should not be used as a substitute for human judgment.

\subsection{Prompts in {\modelname} Benchmark}

In this part, we report the main prompt templates used during {\modelname} construction and evaluation. 

\begin{figure*}[h]
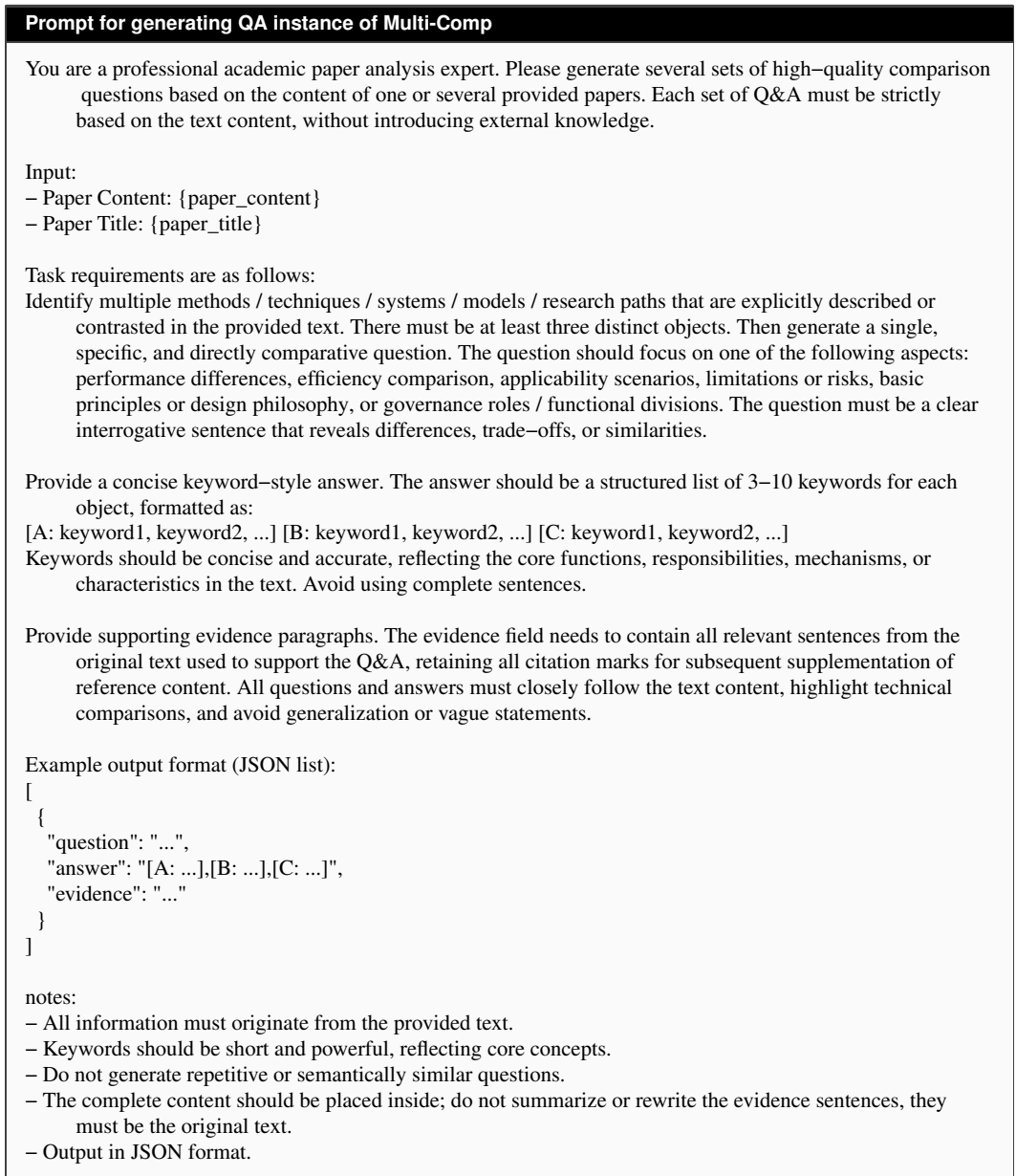

\begin{PromptBox}{Prompt for generating QA instance of Multi-Comp}
You are a professional academic paper analysis expert. Please generate several sets of high-quality comparison questions based on the content of one or several provided papers. Each set of Q&A must be strictly based on the text content, without introducing external knowledge.

Input:
- Paper Content: {paper_content}
- Paper Title: {paper_title}

Task requirements are as follows:
Identify multiple methods / techniques / systems / models / research paths that are explicitly described or contrasted in the provided text. There must be at least three distinct objects. Then generate a single, specific, and directly comparative question. The question should focus on one of the following aspects: performance differences, efficiency comparison, applicability scenarios, limitations or risks, basic principles or design philosophy, or governance roles / functional divisions. The question must be a clear interrogative sentence that reveals differences, trade-offs, or similarities.

Provide a concise keyword-style answer. The answer should be a structured list of 3-10 keywords for each object, formatted as:
[A: keyword1, keyword2, ...] [B: keyword1, keyword2, ...] [C: keyword1, keyword2, ...]
Keywords should be concise and accurate, reflecting the core functions, responsibilities, mechanisms, or characteristics in the text. Avoid using complete sentences.

Provide supporting evidence paragraphs. The evidence field needs to contain all relevant sentences from the original text used to support the Q&A, retaining all citation marks for subsequent supplementation of reference content. All questions and answers must closely follow the text content, highlight technical comparisons, and avoid generalization or vague statements.

Example output format (JSON list):
[
  {
    "question": "...",
    "answer": "[A: ...],[B: ...],[C: ...]",
    "evidence": "..."
  }
]

notes:
- All information must originate from the provided text.
- Keywords should be short and powerful, reflecting core concepts.
- Do not generate repetitive or semantically similar questions.
- The complete content should be placed inside; do not summarize or rewrite the evidence sentences, they must be the original text.
- Output in JSON format.
\end{PromptBox}
\caption{Prompt for generating Multi-Comp QA instance.}
\label{fig:prompt_multi_comparison}
\end{figure*}

\begin{figure*}[h]
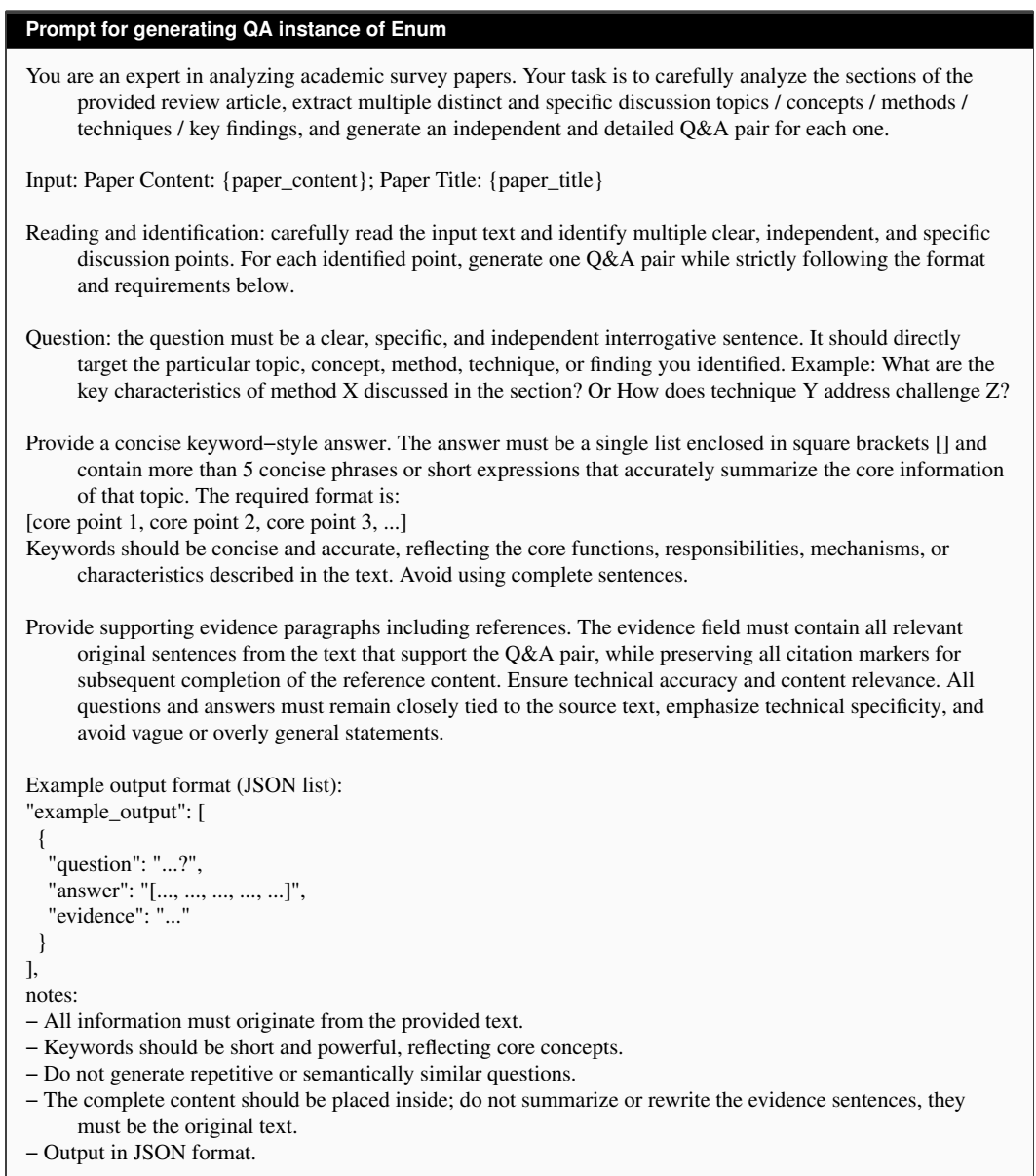

\begin{PromptBox}{Prompt for generating QA instance of Enum}
You are an expert in analyzing academic survey papers. Your task is to carefully analyze the sections of the provided review article, extract multiple distinct and specific discussion topics / concepts / methods / techniques / key findings, and generate an independent and detailed Q&A pair for each one.

Input: Paper Content: {paper_content}; Paper Title: {paper_title}

Reading and identification: carefully read the input text and identify multiple clear, independent, and specific discussion points. For each identified point, generate one Q&A pair while strictly following the format and requirements below.

Question: the question must be a clear, specific, and independent interrogative sentence. It should directly target the particular topic, concept, method, technique, or finding you identified. Example: What are the key characteristics of method X discussed in the section? Or How does technique Y address challenge Z?

Provide a concise keyword-style answer. The answer must be a single list enclosed in square brackets [] and contain more than 5 concise phrases or short expressions that accurately summarize the core information of that topic. The required format is:
[core point 1, core point 2, core point 3, ...]
Keywords should be concise and accurate, reflecting the core functions, responsibilities, mechanisms, or characteristics described in the text. Avoid using complete sentences.

Provide supporting evidence paragraphs including references. The evidence field must contain all relevant original sentences from the text that support the Q&A pair, while preserving all citation markers for subsequent completion of the reference content. Ensure technical accuracy and content relevance. All questions and answers must remain closely tied to the source text, emphasize technical specificity, and avoid vague or overly general statements.

Example output format (JSON list):
"example_output": [
  {
    "question": "...?",
    "answer": "[..., ..., ..., ..., ...]",
    "evidence": "..."
  }
],
notes:
- All information must originate from the provided text.
- Keywords should be short and powerful, reflecting core concepts.
- Do not generate repetitive or semantically similar questions.
- The complete content should be placed inside; do not summarize or rewrite the evidence sentences, they must be the original text.
- Output in JSON format.

\end{PromptBox}
\caption{Prompt for generating QA instance of Enum.}
\label{fig:prompt_gensingle}
\end{figure*}

\begin{figure*}[h]
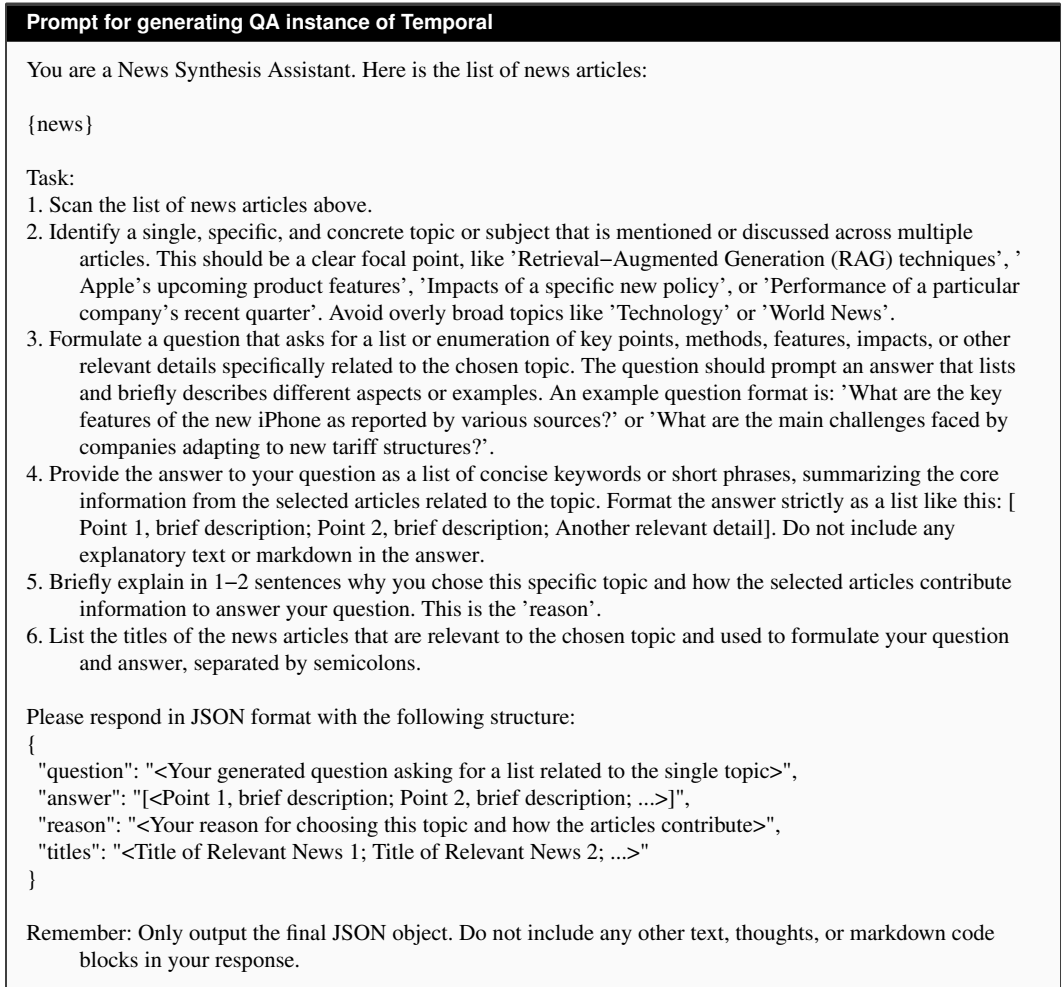

\begin{PromptBox}{Prompt for generating QA instance of Temporal}
You are a News Synthesis Assistant. Here is the list of news articles:

{news}

Task:
1. Scan the list of news articles above.
2. Identify a single, specific, and concrete topic or subject that is mentioned or discussed across multiple articles. This should be a clear focal point, like 'Retrieval-Augmented Generation (RAG) techniques', 'Apple's upcoming product features', 'Impacts of a specific new policy', or 'Performance of a particular company's recent quarter'. Avoid overly broad topics like 'Technology' or 'World News'.
3. Formulate a question that asks for a list or enumeration of key points, methods, features, impacts, or other relevant details specifically related to the chosen topic. The question should prompt an answer that lists and briefly describes different aspects or examples. An example question format is: 'What are the key features of the new iPhone as reported by various sources?' or 'What are the main challenges faced by companies adapting to new tariff structures?'.
4. Provide the answer to your question as a list of concise keywords or short phrases, summarizing the core information from the selected articles related to the topic. Format the answer strictly as a list like this: [Point 1, brief description; Point 2, brief description; Another relevant detail]. Do not include any explanatory text or markdown in the answer.
5. Briefly explain in 1-2 sentences why you chose this specific topic and how the selected articles contribute information to answer your question. This is the 'reason'.
6. List the titles of the news articles that are relevant to the chosen topic and used to formulate your question and answer, separated by semicolons.

Please respond in JSON format with the following structure:
{
  "question": "<Your generated question asking for a list related to the single topic>",
  "answer": "[<Point 1, brief description; Point 2, brief description; ...>]",
  "reason": "<Your reason for choosing this topic and how the articles contribute>",
  "titles": "<Title of Relevant News 1; Title of Relevant News 2; ...>"
}

Remember: Only output the final JSON object. Do not include any other text, thoughts, or markdown code blocks in your response.

\end{PromptBox}
\caption{Prompt for generating Temporal type.}
\label{fig:prompt_temporal_query}
\end{figure*}

We provide prompts for five different question types (\textsc{Single-Sum}, \textsc{Pair-Comp}, \textsc{Multi-Comp}, \textsc{Enumeration}, and \textsc{Temporal}) in Figures~\ref{fig:prompt_single_summarization}, \ref{fig:prompt_double_comparison}, \ref{fig:prompt_multi_comparison}, \ref{fig:prompt_gensingle}, and \ref{fig:prompt_temporal_query}, respectively.

\begin{figure*}[h]
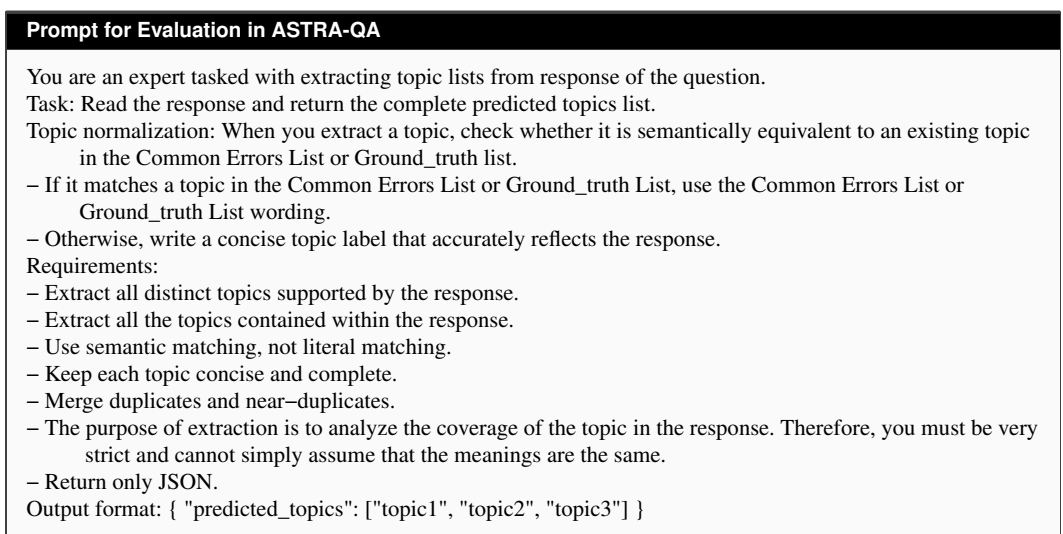

\begin{PromptBox}{Prompt for Evaluation in {\modelname}}
You are an expert tasked with extracting topic lists from response of the question.
Task: Read the response and return the complete predicted topics list.
Topic normalization: When you extract a topic, check whether it is semantically equivalent to an existing topic in the Common Errors List or Ground_truth list.
- If it matches a topic in the Common Errors List or Ground_truth List, use the Common Errors List or Ground_truth List wording.
- Otherwise, write a concise topic label that accurately reflects the response.
Requirements:
- Extract all distinct topics supported by the response.
- Extract all the topics contained within the response.
- Use semantic matching, not literal matching.
- Keep each topic concise and complete.
- Merge duplicates and near-duplicates.
- The purpose of extraction is to analyze the coverage of the topic in the response. Therefore, you must be very strict and cannot simply assume that the meanings are the same.
- Return only JSON.
Output format: { "predicted_topics": ["topic1", "topic2", "topic3"] }
\end{PromptBox}
\caption{Prompt for {\modelname} evaluation method.}
\label{fig:prompt_eval_adc}
\end{figure*}

Additionally, we provide the complete prompt templates employed in {\modelname} and the head-to-head evaluation method, facilitating reproducibility and further analysis in Figure~\ref{fig:prompt_eval_adc}, \ref{fig:prompt_eval_h2h}, respectively.

\section{Experimental Details}

\subsection{Details of Experimental Settings}

\paragraph{Compute resources.} 
All experiments were conducted on a Linux operating system running on a high-performance server equipped with an Intel Xeon 2.0GHz CPU, 1024GB of memory, and 8 NVIDIA GeForce RTX A5000 GPUs, each with 24GB of VRAM. 
Following the experimental setup in Section~\ref{sec:exp}, all RAG methods use Qwen3-8B as the LLM backbone and nomic-embed-text as the embedding model. 
We deploy Qwen3-8B with the vLLM~\cite{kwon2023efficient} framework and serve nomic-embed-text with Ollama~\cite{ollama2024}. 
GPT-5.1 is used for evaluation through API calls. We did not separately log the total wall-clock time and API-token usage for the complete benchmark evaluation. 
However, the total evaluation time can be estimated from the validation-subset comparison in Section~\ref{sec:exp}: on 200 questions, our reference-based evaluation required 7,473.98 seconds and 1,687,701 tokens, compared with 154,784.34 seconds and 36,623,666 tokens for bidirectional head-to-head evaluation, corresponding to about a 95.2\% reduction in evaluation time and a 95.4\% reduction in token usage.

\subsection{Metrics of Dataset Quality for QA}
\label{sec:app-qa-quality-metrics}

We evaluate the quality of QA data in each dataset using five metrics, all computed on a fixed random sample of 500 QA instances per dataset for fair comparison. Let $\mathcal{D}=\{(q_i,a_i)\}_{i=1}^{N}$ denote the sampled QA set, where $N=500$. Following LiveRAG~\cite{carmel2025liverag}, all embedding-based metrics use the SentenceTransformer model \texttt{sentence-transformers/all-MiniLM-L6-v2}. Specifically, we report \textbf{Q\&A-NGD}, \textbf{A-HS}, \textbf{Q-Length}, \textbf{A-Length}, and \textbf{A-CE.}

\begin{itemize}
    \item \textbf{Q\&A-NGD:} Q\&A-NGD (question and answer n-gram diversity) measures lexical diversity on the concatenated question-answer text. A higher value indicates richer lexical variation.
    For each sample, we concatenate the question and answer into $x_i=[q_i;a_i]$. Let $\mathcal{G}_n(\mathcal{D})$ be the multiset of all $n$-grams extracted from $\{x_i\}_{i=1}^{N}$, and let $\mathcal{U}_n(\mathcal{D})$ be the corresponding set of unique $n$-grams. We define
    \[
    \text{Q\&A-NGD}(\mathcal{D})=\sum_{n=1}^{4}\frac{|\mathcal{U}_n(\mathcal{D})|}{|\mathcal{G}_n(\mathcal{D})|}.
    \]

    \item \textbf{A-HS:} A-HS (answer embedding homogeneity score) measures within-answer semantic consistency. For each answer $a_i$, we first split it into a set of answer units $\{u_{i,j}\}_{j=1}^{m_i}$. In practice, if structured claim annotations are available, we use these atomic claims as answer units; otherwise, for free-form answers, we first split the text by line breaks and then further segment each block into sentence-like units using sentence-final punctuation. For list-style answers, each non-empty item is treated as one unit. We then encode each unit as
    \[
    \mathbf{z}_{i,j}=f(u_{i,j}),
    \]
    where $f(\cdot)$ denotes the shared embedding model introduced above. The homogeneity score of answer $a_i$ is defined as the mean pairwise cosine similarity among all unit embeddings:
    \[
    s_i=
    \dfrac{2}{m_i(m_i-1)}\displaystyle\sum_{1\le j<k\le m_i}\cos(\mathbf{z}_{i,j},\mathbf{z}_{i,k}).
    \]
    For answers with fewer than two units, the score is set to $0$ in implementation. The final dataset-level metric is the mean score over all sampled QA pairs:
    \[
    \text{A-HS}(\mathcal{D})=\frac{1}{N}\sum_{i=1}^{N}s_i.
    \]
    A lower value indicates lower semantic homogeneity within an answer, suggesting that the answer tends to cover more diverse aspects and broader topical scope, and is therefore generally more comprehensive.

    \item \textbf{Q-Length:} Question length is the average number of words per question.

    \item \textbf{A-Length:} Answer length is the average number of words in the answer.

    \item \textbf{A-CE:} A-CE (answer cluster entropy) measures the diversity of answer topics in a dataset. We collect all answer units from the sampled QA pairs, embed them using the same model $f(\cdot)$, and cluster them independently for each dataset using KMeans with $K=50$. Let $\{u_t\}_{t=1}^{M}$ be all answer units in the sampled set, let $c_t\in\{1,\dots,K\}$ be the cluster assignment of unit $u_t$, and let
    \[
    p_k=\frac{1}{M}\sum_{t=1}^{M}\mathbf{1}[c_t=k]
    \]
    denote the empirical proportion of cluster $k$. We then compute the Shannon entropy
    \[
    \text{A-CE}(\mathcal{D})=-\sum_{k=1}^{K}p_k\log p_k, \qquad K=50.
    \]
    Here $\log$ denotes the natural logarithm. A higher value indicates broader topical coverage and a more balanced distribution over answer topics.
\end{itemize}

\subsection{Complete Head-to-Head Results}
\label{sec:app-h2h-results}

\newcommand{\apphhcell}[1]{%
  \cellcolor{mycolor1!#1!mycolor2}%
    \raisebox{-0.25ex}{%
      \ifnum#1>49%
        {\normalsize\textcolor{white}{#1}}%
      \else%
        {\normalsize\textcolor{black}{#1}}%
      \fi%
    }%
}

\newsavebox{\apphhbox}
\newenvironment{apphhmatrix}{%
  \centering
  \setlength{\tabcolsep}{5pt}%
  \renewcommand{\arraystretch}{1.5}%
  \begin{lrbox}{\apphhbox}%
}{%
  \end{lrbox}%
  \resizebox{\linewidth}{!}{\usebox{\apphhbox}}%
}

\newcommand{\apphhheader}{%
  & VR & LL & LH & HY & GG & GL & HR & RA & AR & KR & HI \\
}

We report the complete head-to-head results across four evaluation dimensions: 
Comprehensiveness, Diversity, Empowerment, and the Overall Winner selected based on these three criteria.
Each matrix entry is a percentage and always reports the win rate of the row method against the column method. 
For example, in Figure~\ref{fig:app-h2h-full}, the value 92 in the first data column (VR) of the second data row (LL) indicates that LLightRAG (LL) wins 92\% of its pairwise comparisons against Vanilla RAG (VR) under the Overall Winner criterion in the forward-order setting.
The abbreviations VR, LL, LH, HY, GG, GL, HR, RA, AR, KR, and HI denote Vanilla RAG, LLightRAG, HiLightRAG, HyLightRAG, GGraphRAG, LGraphRAG, HippoRAG, RAPTOR, ArchRAG, KET-RAG, and HiRAG, respectively.

\begin{figure*}[h]
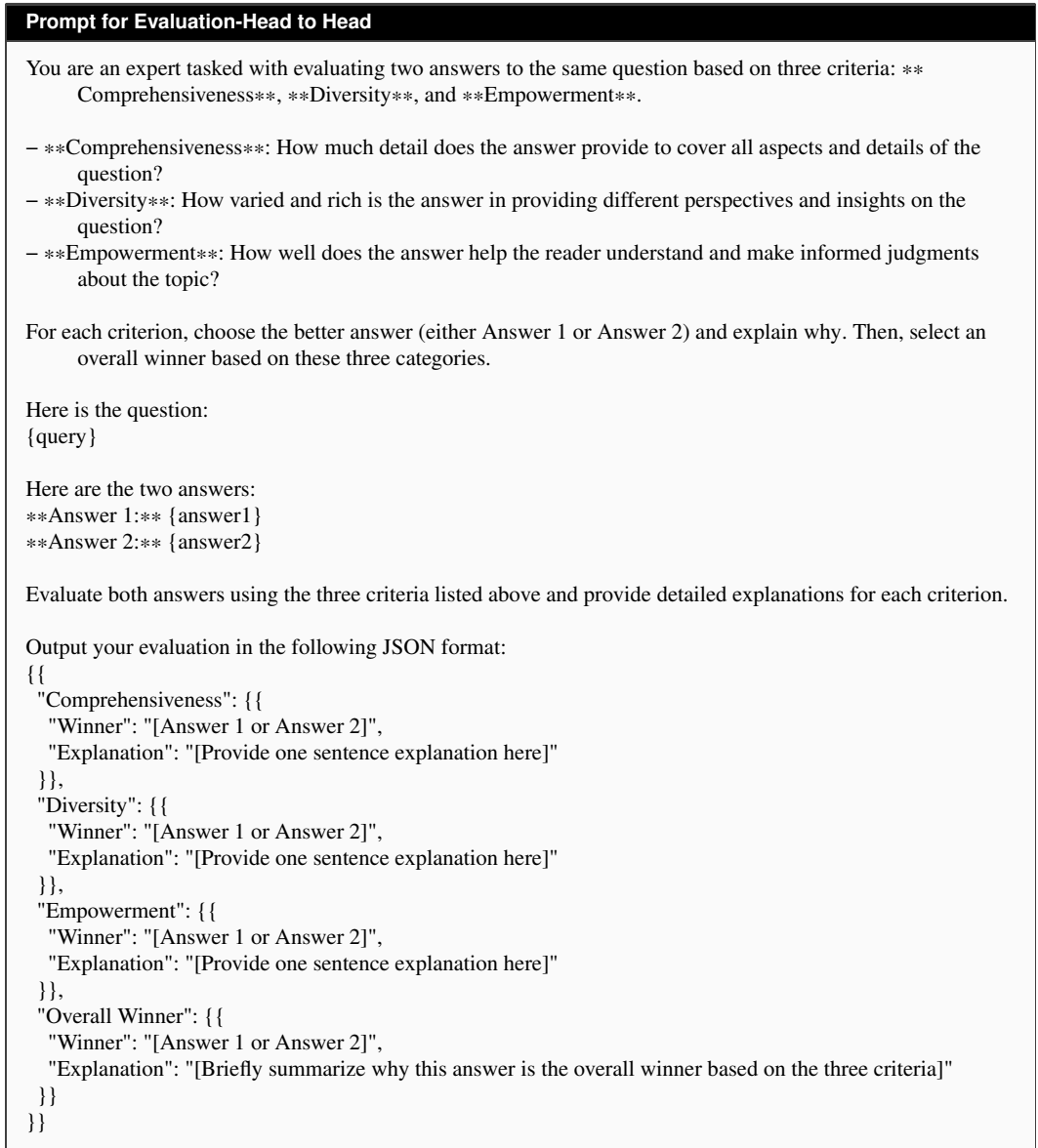

\begin{PromptBox}{Prompt for Evaluation-Head to Head}
You are an expert tasked with evaluating two answers to the same question based on three criteria: **Comprehensiveness**, **Diversity**, and **Empowerment**.

- **Comprehensiveness**: How much detail does the answer provide to cover all aspects and details of the question?
- **Diversity**: How varied and rich is the answer in providing different perspectives and insights on the question?
- **Empowerment**: How well does the answer help the reader understand and make informed judgments about the topic?

For each criterion, choose the better answer (either Answer 1 or Answer 2) and explain why. Then, select an overall winner based on these three categories.

Here is the question:
{query}

Here are the two answers:
**Answer 1:** {answer1}
**Answer 2:** {answer2}

Evaluate both answers using the three criteria listed above and provide detailed explanations for each criterion.

Output your evaluation in the following JSON format:
{{
  "Comprehensiveness": {{
    "Winner": "[Answer 1 or Answer 2]",
    "Explanation": "[Provide one sentence explanation here]"
  }},
  "Diversity": {{
    "Winner": "[Answer 1 or Answer 2]",
    "Explanation": "[Provide one sentence explanation here]"
  }},
  "Empowerment": {{
    "Winner": "[Answer 1 or Answer 2]",
    "Explanation": "[Provide one sentence explanation here]"
  }},
  "Overall Winner": {{
    "Winner": "[Answer 1 or Answer 2]",
    "Explanation": "[Briefly summarize why this answer is the overall winner based on the three criteria]"
  }}
}}
\end{PromptBox}
\caption{Prompt for head-to-head evaluation.}
\label{fig:prompt_eval_h2h}
\end{figure*}

Following the head-to-head evaluation prompt in Figure~\ref{fig:prompt_eval_h2h}, the \textbf{forward order} places the row method's output in the Answer 1 position and the column method's output in the Answer 2 position.
The \textbf{reverse order} places the column method's output in the Answer 1 position and the row method's output in the Answer 2 position.
These two settings can lead to different win rates because the presentation order may affect the LLM-based head-to-head judgment.
For example, for the LL-vs-LH comparison under the Overall Winner criterion, LL obtains a 33\% win rate when LL is shown before LH in Figure~\ref{fig:app-h2h-full} (second row, third column), but a 47\% win rate when LH is shown before LL in Figure~\ref{fig:app-h2h-column-win} (second row, third column). 

\begin{figure*}[ht]
    \centering
\begin{subfigure}[t]{0.49\textwidth}
\begin{apphhmatrix}
\begin{tabular}{c*{11}{c}}
\apphhheader
VR  & \apphhcell{50} & \apphhcell{8}  & \apphhcell{5}  & \apphhcell{36} & \apphhcell{41} & \apphhcell{17} & \apphhcell{3}  & \apphhcell{1}  & \apphhcell{1} & \apphhcell{91} & \apphhcell{14} \\
LL  & \apphhcell{92} & \apphhcell{50} & \apphhcell{33} & \apphhcell{60} & \apphhcell{86} & \apphhcell{44} & \apphhcell{29} & \apphhcell{14} & \apphhcell{23} & \apphhcell{95} & \apphhcell{78} \\
LH & \apphhcell{95} & \apphhcell{67} & \apphhcell{50} & \apphhcell{60} & \apphhcell{90} & \apphhcell{52} & \apphhcell{37} & \apphhcell{26} & \apphhcell{31} & \apphhcell{98} & \apphhcell{82} \\
HY & \apphhcell{64} & \apphhcell{40} & \apphhcell{40} & \apphhcell{50} & \apphhcell{60} & \apphhcell{23} & \apphhcell{28} & \apphhcell{11} & \apphhcell{10} & \apphhcell{88} & \apphhcell{44} \\
GG  & \apphhcell{59} & \apphhcell{14} & \apphhcell{10} & \apphhcell{40} & \apphhcell{50} & \apphhcell{21} & \apphhcell{19} & \apphhcell{2}  & \apphhcell{1} & \apphhcell{73} & \apphhcell{30} \\
GL  & \apphhcell{83} & \apphhcell{56} & \apphhcell{48} & \apphhcell{77} & \apphhcell{79} & \apphhcell{50} & \apphhcell{35} & \apphhcell{21} & \apphhcell{28} & \apphhcell{95} & \apphhcell{71} \\
HR  & \apphhcell{97} & \apphhcell{71} & \apphhcell{63} & \apphhcell{72} & \apphhcell{81} & \apphhcell{65} & \apphhcell{50} & \apphhcell{40} & \apphhcell{49} & \apphhcell{98} & \apphhcell{96} \\
RA  & \apphhcell{99} & \apphhcell{86} & \apphhcell{74} & \apphhcell{89} & \apphhcell{98} & \apphhcell{79} & \apphhcell{60} & \apphhcell{50} & \apphhcell{54} & \apphhcell{99} & \apphhcell{96} \\
AR  & \apphhcell{99} & \apphhcell{77} & \apphhcell{69} & \apphhcell{90} & \apphhcell{99} & \apphhcell{72} & \apphhcell{51} & \apphhcell{46} & \apphhcell{50} & \apphhcell{99} & \apphhcell{99} \\
KR  & \apphhcell{9}  & \apphhcell{5}  & \apphhcell{2}  & \apphhcell{12} & \apphhcell{27} & \apphhcell{5}  & \apphhcell{2}  & \apphhcell{1}  & \apphhcell{1} & \apphhcell{50} & \apphhcell{1}  \\
HI  & \apphhcell{86} & \apphhcell{22} & \apphhcell{18} & \apphhcell{56} & \apphhcell{70} & \apphhcell{29} & \apphhcell{4}  & \apphhcell{4}  & \apphhcell{1} & \apphhcell{99} & \apphhcell{50} \\

\end{tabular}
\end{apphhmatrix}
    \caption{Overall Winner, forward order}
    \label{fig:app-h2h-full}
\end{subfigure}
\hfill
\begin{subfigure}[t]{0.49\textwidth}
\begin{apphhmatrix}
\begin{tabular}{c*{11}{c}}
\apphhheader
VR  & \apphhcell{50} & \apphhcell{11} & \apphhcell{6}  & \apphhcell{44} & \apphhcell{49} & \apphhcell{22} & \apphhcell{4}  & \apphhcell{1}  & \apphhcell{1}  & \apphhcell{92} & \apphhcell{19} \\
LL  & \apphhcell{89} & \apphhcell{50} & \apphhcell{47} & \apphhcell{69} & \apphhcell{90} & \apphhcell{55} & \apphhcell{29} & \apphhcell{27} & \apphhcell{30} & \apphhcell{98} & \apphhcell{85} \\
LH  & \apphhcell{94} & \apphhcell{53} & \apphhcell{50} & \apphhcell{71} & \apphhcell{92} & \apphhcell{57} & \apphhcell{41} & \apphhcell{37} & \apphhcell{40} & \apphhcell{98} & \apphhcell{86} \\
HY  & \apphhcell{56} & \apphhcell{31} & \apphhcell{29} & \apphhcell{50} & \apphhcell{61} & \apphhcell{38} & \apphhcell{20} & \apphhcell{14} & \apphhcell{13} & \apphhcell{88} & \apphhcell{49} \\
GG  & \apphhcell{51} & \apphhcell{10} & \apphhcell{8}  & \apphhcell{39} & \apphhcell{50} & \apphhcell{20} & \apphhcell{4}  & \apphhcell{1}  & \apphhcell{2}  & \apphhcell{73} & \apphhcell{41} \\
GL  & \apphhcell{78} & \apphhcell{45} & \apphhcell{43} & \apphhcell{62} & \apphhcell{80} & \apphhcell{50} & \apphhcell{33} & \apphhcell{30} & \apphhcell{31} & \apphhcell{95} & \apphhcell{72} \\
HR  & \apphhcell{96} & \apphhcell{71} & \apphhcell{59} & \apphhcell{80} & \apphhcell{96} & \apphhcell{67} & \apphhcell{50} & \apphhcell{54} & \apphhcell{55} & \apphhcell{98} & \apphhcell{6}  \\
RA  & \apphhcell{99} & \apphhcell{73} & \apphhcell{63} & \apphhcell{86} & \apphhcell{99} & \apphhcell{70} & \apphhcell{46} & \apphhcell{50} & \apphhcell{57} & \apphhcell{99} & \apphhcell{6}  \\
AR  & \apphhcell{99} & \apphhcell{70} & \apphhcell{60} & \apphhcell{87} & \apphhcell{98} & \apphhcell{69} & \apphhcell{45} & \apphhcell{43} & \apphhcell{50} & \apphhcell{99} & \apphhcell{99} \\
KR  & \apphhcell{8}  & \apphhcell{2}  & \apphhcell{2}  & \apphhcell{12} & \apphhcell{27} & \apphhcell{5}  & \apphhcell{2}  & \apphhcell{1}  & \apphhcell{1}  & \apphhcell{50} & \apphhcell{2}  \\
HI  & \apphhcell{81} & \apphhcell{15} & \apphhcell{14} & \apphhcell{51} & \apphhcell{59} & \apphhcell{28} & \apphhcell{4}  & \apphhcell{4}  & \apphhcell{1}  & \apphhcell{98} & \apphhcell{50} \\

\end{tabular}
\end{apphhmatrix}
    \caption{Overall Winner, reverse order}

    \label{fig:app-h2h-column-win}
\end{subfigure}
    \caption{Head-to-head win rates for the \textbf{Overall Winner} criterion. Each entry reports the row method's win rate against the column method; higher is better.}
\end{figure*}

\begin{figure*}[ht]
    \centering
\begin{subfigure}[t]{0.49\textwidth}
\begin{apphhmatrix}
\begin{tabular}{c*{11}{c}}
\apphhheader
VR  & \apphhcell{50} & \apphhcell{9}  & \apphhcell{6}  & \apphhcell{37} & \apphhcell{41} & \apphhcell{17} & \apphhcell{3}  & \apphhcell{2}  & \apphhcell{1} & \apphhcell{90} & \apphhcell{14} \\
LL  & \apphhcell{91} & \apphhcell{50} & \apphhcell{36} & \apphhcell{60} & \apphhcell{89} & \apphhcell{45} & \apphhcell{23} & \apphhcell{19} & \apphhcell{23} & \apphhcell{96} & \apphhcell{77} \\
LH & \apphhcell{94} & \apphhcell{64} & \apphhcell{50} & \apphhcell{62} & \apphhcell{89} & \apphhcell{52} & \apphhcell{30} & \apphhcell{29} & \apphhcell{29} & \apphhcell{97} & \apphhcell{81} \\
HY & \apphhcell{63} & \apphhcell{40} & \apphhcell{38} & \apphhcell{50} & \apphhcell{61} & \apphhcell{31} & \apphhcell{17} & \apphhcell{13} & \apphhcell{8} & \apphhcell{91} & \apphhcell{45} \\
GG  & \apphhcell{59} & \apphhcell{11} & \apphhcell{11} & \apphhcell{39} & \apphhcell{50} & \apphhcell{19} & \apphhcell{5}  & \apphhcell{4}  & \apphhcell{1} & \apphhcell{71} & \apphhcell{31} \\
GL  & \apphhcell{83} & \apphhcell{55} & \apphhcell{48} & \apphhcell{69} & \apphhcell{81} & \apphhcell{50} & \apphhcell{29} & \apphhcell{25} & \apphhcell{26} & \apphhcell{96} & \apphhcell{72} \\
HR  & \apphhcell{97} & \apphhcell{77} & \apphhcell{70} & \apphhcell{83} & \apphhcell{95} & \apphhcell{71} & \apphhcell{50} & \apphhcell{47} & \apphhcell{46} & \apphhcell{99} & \apphhcell{94} \\
RA  & \apphhcell{98} & \apphhcell{81} & \apphhcell{71} & \apphhcell{87} & \apphhcell{96} & \apphhcell{75} & \apphhcell{53} & \apphhcell{50} & \apphhcell{50} & \apphhcell{99} & \apphhcell{92} \\
AR  & \apphhcell{99} & \apphhcell{77} & \apphhcell{71} & \apphhcell{92} & \apphhcell{99} & \apphhcell{74} & \apphhcell{54} & \apphhcell{50} & \apphhcell{50} &  \apphhcell{99} & \apphhcell{99} \\
KR  & \apphhcell{10} & \apphhcell{4}  & \apphhcell{3}  & \apphhcell{9}  & \apphhcell{29} & \apphhcell{4}  & \apphhcell{1}  & \apphhcell{1}  & \apphhcell{1} & \apphhcell{50} & \apphhcell{1}  \\
HI  & \apphhcell{86} & \apphhcell{23} & \apphhcell{19} & \apphhcell{55} & \apphhcell{69} & \apphhcell{28} & \apphhcell{6}  & \apphhcell{8}  & \apphhcell{1} & \apphhcell{99} & \apphhcell{50} \\

\end{tabular}
\end{apphhmatrix}
    \caption{Comprehensiveness, forward order}
    \label{fig:app-h2h-comp-row-win}
\end{subfigure}
\hfill
\begin{subfigure}[t]{0.49\textwidth}
\begin{apphhmatrix}
\begin{tabular}{c*{11}{c}}
\apphhheader
VR  & \apphhcell{50} & \apphhcell{10} & \apphhcell{5}  & \apphhcell{38} & \apphhcell{39} & \apphhcell{18} & \apphhcell{3}  & \apphhcell{2}  & \apphhcell{1}  & \apphhcell{87} & \apphhcell{13} \\
LL  & \apphhcell{90} & \apphhcell{50} & \apphhcell{42} & \apphhcell{66} & \apphhcell{89} & \apphhcell{52} & \apphhcell{29} & \apphhcell{26} & \apphhcell{23} & \apphhcell{97} & \apphhcell{83} \\
LH  & \apphhcell{95} & \apphhcell{58} & \apphhcell{50} & \apphhcell{67} & \apphhcell{90} & \apphhcell{53} & \apphhcell{37} & \apphhcell{35} & \apphhcell{32} & \apphhcell{98} & \apphhcell{84} \\
HY  & \apphhcell{62} & \apphhcell{34} & \apphhcell{33} & \apphhcell{50} & \apphhcell{60} & \apphhcell{38} & \apphhcell{19} & \apphhcell{15} & \apphhcell{10} & \apphhcell{88} & \apphhcell{46} \\
GG  & \apphhcell{61} & \apphhcell{11} & \apphhcell{10} & \apphhcell{40} & \apphhcell{50} & \apphhcell{17} & \apphhcell{4}  & \apphhcell{3}  & \apphhcell{1}  & \apphhcell{72} & \apphhcell{35} \\
GL  & \apphhcell{82} & \apphhcell{48} & \apphhcell{47} & \apphhcell{62} & \apphhcell{83} & \apphhcell{50} & \apphhcell{33} & \apphhcell{28} & \apphhcell{23} & \apphhcell{96} & \apphhcell{73} \\
HR  & \apphhcell{97} & \apphhcell{71} & \apphhcell{63} & \apphhcell{81} & \apphhcell{96} & \apphhcell{67} & \apphhcell{50} & \apphhcell{50} & \apphhcell{43} & \apphhcell{98} & \apphhcell{93} \\
RA  & \apphhcell{98} & \apphhcell{74} & \apphhcell{65} & \apphhcell{85} & \apphhcell{97} & \apphhcell{72} & \apphhcell{50} & \apphhcell{50} & \apphhcell{46} & \apphhcell{99} & \apphhcell{92} \\
AR  & \apphhcell{99} & \apphhcell{77} & \apphhcell{68} & \apphhcell{90} & \apphhcell{99} & \apphhcell{77} & \apphhcell{57} & \apphhcell{54} & \apphhcell{50} & \apphhcell{99} & \apphhcell{99} \\
KR  & \apphhcell{13} & \apphhcell{3}  & \apphhcell{2}  & \apphhcell{12} & \apphhcell{28} & \apphhcell{4}  & \apphhcell{2}  & \apphhcell{1}  & \apphhcell{1}  & \apphhcell{50} & \apphhcell{1}  \\
HI  & \apphhcell{87} & \apphhcell{17} & \apphhcell{16} & \apphhcell{54} & \apphhcell{65} & \apphhcell{27} & \apphhcell{7}  & \apphhcell{8}  & \apphhcell{1}  & \apphhcell{99} & \apphhcell{50} \\
\end{tabular}
\end{apphhmatrix}
    \caption{Comprehensiveness, reverse order}
    \label{fig:app-h2h-comp-column-win}
\end{subfigure}
    \caption{Head-to-head win rates for the \textbf{Comprehensiveness} criterion. Each entry reports the row method's win rate against the column method; higher is better.}
\end{figure*}

\begin{figure*}[ht]
    \centering
\begin{subfigure}[t]{0.49\textwidth}
\begin{apphhmatrix}
\begin{tabular}{c*{11}{c}}
\apphhheader
VR  & \apphhcell{50} & \apphhcell{8}  & \apphhcell{6}  & \apphhcell{35} & \apphhcell{38} & \apphhcell{15} & \apphhcell{2}  & \apphhcell{2}  & \apphhcell{1} & \apphhcell{88} & \apphhcell{10} \\
LL  & \apphhcell{92} & \apphhcell{50} & \apphhcell{36} & \apphhcell{57} & \apphhcell{89} & \apphhcell{42} & \apphhcell{23} & \apphhcell{18} & \apphhcell{10} & \apphhcell{96} & \apphhcell{77} \\
LH & \apphhcell{94} & \apphhcell{64} & \apphhcell{50} & \apphhcell{59} & \apphhcell{89} & \apphhcell{48} & \apphhcell{27} & \apphhcell{27} & \apphhcell{13} & \apphhcell{96} & \apphhcell{81} \\
HY & \apphhcell{65} & \apphhcell{43} & \apphhcell{41} & \apphhcell{50} & \apphhcell{62} & \apphhcell{28} & \apphhcell{16} & \apphhcell{13} & \apphhcell{4} & \apphhcell{92} & \apphhcell{44} \\
GG  & \apphhcell{62} & \apphhcell{11} & \apphhcell{11} & \apphhcell{38} & \apphhcell{50} & \apphhcell{19} & \apphhcell{5}  & \apphhcell{4}  & \apphhcell{1} & \apphhcell{71} & \apphhcell{29} \\
GL  & \apphhcell{85} & \apphhcell{58} & \apphhcell{52} & \apphhcell{72} & \apphhcell{81} & \apphhcell{50} & \apphhcell{27} & \apphhcell{23} & \apphhcell{12} & \apphhcell{96} & \apphhcell{72} \\
HR  & \apphhcell{98} & \apphhcell{77} & \apphhcell{73} & \apphhcell{84} & \apphhcell{95} & \apphhcell{73} & \apphhcell{50} & \apphhcell{43} & \apphhcell{24} & \apphhcell{99} & \apphhcell{93} \\
RA  & \apphhcell{98} & \apphhcell{82} & \apphhcell{73} & \apphhcell{87} & \apphhcell{96} & \apphhcell{77} & \apphhcell{57} & \apphhcell{50} & \apphhcell{24} & \apphhcell{99} & \apphhcell{92} \\
AR  & \apphhcell{99} & \apphhcell{90} & \apphhcell{87} & \apphhcell{96} & \apphhcell{99} & \apphhcell{88} & \apphhcell{76} & \apphhcell{76} & \apphhcell{50} & \apphhcell{99} & \apphhcell{99} \\
KR  & \apphhcell{12} & \apphhcell{4}  & \apphhcell{4}  & \apphhcell{8}  & \apphhcell{29} & \apphhcell{4}  & \apphhcell{1}  & \apphhcell{1}  & \apphhcell{1} & \apphhcell{50} & \apphhcell{1}  \\
HI  & \apphhcell{90} & \apphhcell{23} & \apphhcell{19} & \apphhcell{56} & \apphhcell{71} & \apphhcell{28} & \apphhcell{7}  & \apphhcell{8}  & \apphhcell{1} & \apphhcell{99} & \apphhcell{50} \\

\end{tabular}
\end{apphhmatrix}
    \caption{Diversity, forward order}
    \label{tab:app-h2h-div-row-win}
\end{subfigure}
\hfill
\begin{subfigure}[t]{0.49\textwidth}
\begin{apphhmatrix}
\begin{tabular}{c*{11}{c}}
\apphhheader
VR  & \apphhcell{50} & \apphhcell{10} & \apphhcell{6}  & \apphhcell{38} & \apphhcell{39} & \apphhcell{18} & \apphhcell{2}  & \apphhcell{2}  & \apphhcell{1}  & \apphhcell{89} & \apphhcell{17} \\
LL  & \apphhcell{90} & \apphhcell{50} & \apphhcell{46} & \apphhcell{66} & \apphhcell{90} & \apphhcell{55} & \apphhcell{34} & \apphhcell{33} & \apphhcell{20} & \apphhcell{96} & \apphhcell{86} \\
LH  & \apphhcell{94} & \apphhcell{54} & \apphhcell{50} & \apphhcell{66} & \apphhcell{90} & \apphhcell{58} & \apphhcell{39} & \apphhcell{41} & \apphhcell{26} & \apphhcell{97} & \apphhcell{86} \\
HY  & \apphhcell{62} & \apphhcell{34} & \apphhcell{34} & \apphhcell{50} & \apphhcell{61} & \apphhcell{41} & \apphhcell{21} & \apphhcell{20} & \apphhcell{9}  & \apphhcell{90} & \apphhcell{47} \\
GG  & \apphhcell{61} & \apphhcell{10} & \apphhcell{10} & \apphhcell{39} & \apphhcell{50} & \apphhcell{18} & \apphhcell{6}  & \apphhcell{4}  & \apphhcell{1}  & \apphhcell{72} & \apphhcell{39} \\
GL  & \apphhcell{82} & \apphhcell{45} & \apphhcell{42} & \apphhcell{59} & \apphhcell{82} & \apphhcell{50} & \apphhcell{38} & \apphhcell{34} & \apphhcell{23} & \apphhcell{97} & \apphhcell{72} \\
HR  & \apphhcell{98} & \apphhcell{66} & \apphhcell{61} & \apphhcell{79} & \apphhcell{94} & \apphhcell{62} & \apphhcell{50} & \apphhcell{53} & \apphhcell{39} & \apphhcell{99} & \apphhcell{94} \\
RA  & \apphhcell{98} & \apphhcell{67} & \apphhcell{59} & \apphhcell{80} & \apphhcell{96} & \apphhcell{66} & \apphhcell{47} & \apphhcell{50} & \apphhcell{39} & \apphhcell{99} & \apphhcell{92} \\
AR  & \apphhcell{99} & \apphhcell{80} & \apphhcell{74} & \apphhcell{91} & \apphhcell{99} & \apphhcell{77} & \apphhcell{61} & \apphhcell{61} & \apphhcell{50} & \apphhcell{99} & \apphhcell{99} \\
KR  & \apphhcell{11} & \apphhcell{4}  & \apphhcell{3}  & \apphhcell{10} & \apphhcell{28} & \apphhcell{3}  & \apphhcell{1}  & \apphhcell{1}  & \apphhcell{1}  & \apphhcell{50} & \apphhcell{2}  \\
HI  & \apphhcell{83} & \apphhcell{14} & \apphhcell{14} & \apphhcell{53} & \apphhcell{61} & \apphhcell{28} & \apphhcell{6}  & \apphhcell{8}  & \apphhcell{1}  & \apphhcell{98} & \apphhcell{50} \\

\end{tabular}
\end{apphhmatrix}
    \caption{Diversity, reverse order}
    \label{tab:app-h2h-div-column-win}
\end{subfigure}
    \caption{Head-to-head win rates for the \textbf{Diversity} criterion. Each entry reports the row method's win rate against the column method; higher is better.}
\end{figure*}

\begin{figure*}[ht]
    \centering
\begin{subfigure}[t]{0.49\textwidth}
\begin{apphhmatrix}
\begin{tabular}{c*{11}{c}}
\apphhheader
VR  & \apphhcell{50} & \apphhcell{7}  & \apphhcell{5}  & \apphhcell{36} & \apphhcell{40} & \apphhcell{17} & \apphhcell{4}  & \apphhcell{1}  & \apphhcell{1} & \apphhcell{91} & \apphhcell{11} \\
LL  & \apphhcell{93} & \apphhcell{50} & \apphhcell{34} & \apphhcell{58} & \apphhcell{88} & \apphhcell{43} & \apphhcell{21} & \apphhcell{17} & \apphhcell{21} & \apphhcell{96} & \apphhcell{77} \\
LH & \apphhcell{95} & \apphhcell{66} & \apphhcell{50} & \apphhcell{60} & \apphhcell{88} & \apphhcell{47} & \apphhcell{27} & \apphhcell{27} & \apphhcell{32} & \apphhcell{97} & \apphhcell{78} \\
HY & \apphhcell{64} & \apphhcell{42} & \apphhcell{40} & \apphhcell{50} & \apphhcell{61} & \apphhcell{26} & \apphhcell{16} & \apphhcell{13} & \apphhcell{10} & \apphhcell{90} & \apphhcell{42} \\
GG  & \apphhcell{60} & \apphhcell{12} & \apphhcell{12} & \apphhcell{39} & \apphhcell{50} & \apphhcell{19} & \apphhcell{5}  & \apphhcell{4}  & \apphhcell{1} & \apphhcell{71} & \apphhcell{29} \\
GL  & \apphhcell{83} & \apphhcell{57} & \apphhcell{53} & \apphhcell{74} & \apphhcell{81} & \apphhcell{50} & \apphhcell{27} & \apphhcell{23} & \apphhcell{28} & \apphhcell{96} & \apphhcell{72} \\
HR  & \apphhcell{96} & \apphhcell{79} & \apphhcell{73} & \apphhcell{84} & \apphhcell{95} & \apphhcell{73} & \apphhcell{50} & \apphhcell{42} & \apphhcell{48} & \apphhcell{99} & \apphhcell{92} \\
RA  & \apphhcell{99} & \apphhcell{83} & \apphhcell{73} & \apphhcell{87} & \apphhcell{96} & \apphhcell{77} & \apphhcell{58} & \apphhcell{50} & \apphhcell{53} & \apphhcell{99} & \apphhcell{91} \\
AR  & \apphhcell{99} & \apphhcell{79} & \apphhcell{68} & \apphhcell{90} & \apphhcell{99} & \apphhcell{72} & \apphhcell{52} & \apphhcell{47} & \apphhcell{50} & \apphhcell{99} & \apphhcell{99} \\
KR  & \apphhcell{9}  & \apphhcell{4}  & \apphhcell{3}  & \apphhcell{10} & \apphhcell{29} & \apphhcell{4}  & \apphhcell{1}  & \apphhcell{1}  & \apphhcell{1} & \apphhcell{50} & \apphhcell{1}  \\
HI  & \apphhcell{89} & \apphhcell{23} & \apphhcell{22} & \apphhcell{58} & \apphhcell{71} & \apphhcell{28} & \apphhcell{8}  & \apphhcell{9}  & \apphhcell{1} & \apphhcell{99} & \apphhcell{50} \\

\end{tabular}
\end{apphhmatrix}
    \caption{Empowerment, forward order}
    \label{tab:app-h2h-emp-row-win}
\end{subfigure}
\hfill
\begin{subfigure}[t]{0.49\textwidth}
\begin{apphhmatrix}
\begin{tabular}{c*{11}{c}}
\apphhheader
VR  & \apphhcell{50} & \apphhcell{10} & \apphhcell{6}  & \apphhcell{38} & \apphhcell{39} & \apphhcell{18} & \apphhcell{2}  & \apphhcell{2}  & \apphhcell{1}  & \apphhcell{89} & \apphhcell{17} \\
LL  & \apphhcell{90} & \apphhcell{50} & \apphhcell{46} & \apphhcell{66} & \apphhcell{90} & \apphhcell{55} & \apphhcell{34} & \apphhcell{33} & \apphhcell{31} & \apphhcell{96} & \apphhcell{86} \\
LH  & \apphhcell{94} & \apphhcell{54} & \apphhcell{50} & \apphhcell{66} & \apphhcell{90} & \apphhcell{58} & \apphhcell{39} & \apphhcell{41} & \apphhcell{40} & \apphhcell{97} & \apphhcell{86} \\
HY  & \apphhcell{62} & \apphhcell{34} & \apphhcell{34} & \apphhcell{50} & \apphhcell{61} & \apphhcell{41} & \apphhcell{21} & \apphhcell{20} & \apphhcell{13} & \apphhcell{90} & \apphhcell{47} \\
GG  & \apphhcell{61} & \apphhcell{10} & \apphhcell{10} & \apphhcell{39} & \apphhcell{50} & \apphhcell{18} & \apphhcell{6}  & \apphhcell{4}  & \apphhcell{2}  & \apphhcell{72} & \apphhcell{39} \\
GL  & \apphhcell{82} & \apphhcell{45} & \apphhcell{42} & \apphhcell{59} & \apphhcell{82} & \apphhcell{50} & \apphhcell{38} & \apphhcell{34} & \apphhcell{31} & \apphhcell{97} & \apphhcell{72} \\
HR  & \apphhcell{98} & \apphhcell{66} & \apphhcell{61} & \apphhcell{79} & \apphhcell{94} & \apphhcell{62} & \apphhcell{50} & \apphhcell{53} & \apphhcell{55} & \apphhcell{99} & \apphhcell{94} \\
RA  & \apphhcell{98} & \apphhcell{67} & \apphhcell{59} & \apphhcell{80} & \apphhcell{96} & \apphhcell{66} & \apphhcell{47} & \apphhcell{50} & \apphhcell{58} & \apphhcell{99} & \apphhcell{92} \\
AR  & \apphhcell{99} & \apphhcell{69} & \apphhcell{60} & \apphhcell{87} & \apphhcell{98} & \apphhcell{69} & \apphhcell{45} & \apphhcell{42} & \apphhcell{50} & \apphhcell{99} & \apphhcell{99} \\
KR  & \apphhcell{11} & \apphhcell{4}  & \apphhcell{3}  & \apphhcell{10} & \apphhcell{28} & \apphhcell{3}  & \apphhcell{1}  & \apphhcell{1}  & \apphhcell{1}  & \apphhcell{50} & \apphhcell{2}  \\
HI  & \apphhcell{83} & \apphhcell{14} & \apphhcell{14} & \apphhcell{53} & \apphhcell{61} & \apphhcell{28} & \apphhcell{6}  & \apphhcell{8}  & \apphhcell{1}  & \apphhcell{98} & \apphhcell{50} \\

\end{tabular}
\end{apphhmatrix}
    \caption{Empowerment, reverse order}
    \label{tab:app-h2h-emp-column-win}
\end{subfigure}
    \caption{Head-to-head win rates for the \textbf{Empowerment} criterion. Each entry reports the row method's win rate against the column method; higher is better.}
\end{figure*}

\subsection{Topic Precision and Recall Results}
\label{sec:app-retrieval-prec-rec}

Table~\ref{tab:retrieval-prec-rec} complements the complete topic precision and topic recall. 
We highlight the best score for each column in {\setlength{\fboxsep}{0.5pt}\colorbox{best}{red}}, with higher T-Prec and T-Rec indicating better performance.

\begin{table*}[ht]
    \centering
    \caption{Topic precision and recall of representative RAG methods on {\modelname} under the three retrieval settings.
    T-Prec and T-Rec denote topic precision and topic recall, respectively.
    }
\small
{
    \begin{tabular}{l|cc|cc|cc}
\toprule
\multirow{2}{*}{Method} & \multicolumn{2}{c}{Simple} & \multicolumn{2}{c}{Middle} & \multicolumn{2}{c}{Hard} \\
\cmidrule(lr){2-3} \cmidrule(lr){4-5} \cmidrule(lr){6-7}
& T-Prec & T-Rec & T-Prec & T-Rec & T-Prec & T-Rec \\
\midrule
{Vanilla RAG} & 55.6 & 42.7 & 39.6 & 19.5 & 39.9 & 18.6 \\
{LLightRAG} & 68.7 & 50.7 & 53.8 & 33.7 & 49.5 & 30.9 \\
{HiLightRAG} & 66.1 & 50.4 & 61.7 & 46.2 & \cellcolor{best}{68.2} & 41.1 \\
{HyLightRAG} & 66.4 & 34.8 & 60.1 & 34.4 & 52.6 & 34.1 \\
{LGraphRAG} & 73.3 & 53.8 & 59.0 & 35.2 & 56.7 & 34.2 \\
{GGraphRAG} & 40.2 & 18.2 & 46.1 & 16.7 & 35.5 & 17.4 \\
{HippoRAG} & 71.1 & 60.1 & 65.3 & \cellcolor{best}{56.8} & 61.9 & 50.6 \\
{RAPTOR} & 71.9 & 63.5 & 63.6 & 54.1 & 64.5 & \cellcolor{best}{50.8} \\
{ArchRAG} & 62.4 & 57.9 & 62.8 & 47.6 & 62.4 & 46.7 \\
{KET-RAG} & 59.5 & 24.1 & 51.6 & 20.4 & 22.1 & 10.1 \\
{HiRAG} & \cellcolor{best}{77.3} & \cellcolor{best}{67.5} & \cellcolor{best}{68.7} & 38.0 & 67.0 & 27.8 \\
\bottomrule
    \end{tabular}
}
    \label{tab:retrieval-prec-rec}
\end{table*}

HiRAG performs best under \textsc{Simple}, reaching 77.3 T-Prec and 67.5 T-Rec, and also obtains the highest \textsc{Middle} T-Prec of 68.7, but its recall drops as the retrieval scope expands. In contrast, HippoRAG and RAPTOR show stronger recall-oriented behavior under broader retrieval settings, achieving the best \textsc{Middle} and \textsc{Hard} T-Rec scores of 56.8 and 50.8, respectively, while HiLightRAG obtains the best \textsc{Hard} T-Prec of 68.2. These results indicate that {\modelname} distinguishes conservative high-precision behavior from broader recall-oriented topic coverage, complementing the results in the main tables.

\begin{figure*}[h]
\begin{PromptBox}{Prompt for generating QA instance of Single-Sum}
Please analyze the provided academic paper reviews and extract more than 10 technical terms or key phrases that best represent the paper's core contributions, methodological focus, and major technical concerns.

Instructions:
1. Read all review content, including summaries, strengths, and weaknesses.
2. Extract technical terms or key phrases that are explicitly mentioned or strongly implied by the reviews.
3. First identify the main topics mentioned in each individual review, then aggregate them across all reviews.
4. Prioritize compound technical phrases over single words.
5. Exclude:
   - general academic verbs or adjectives
   - generic praise or criticism
   - explanations, commentary, or full sentences

Output Requirements:
- Output only technical terms or key phrases
- Do not include any introduction or explanation
- Use a comma-separated list
- Do not use markdown
- Do not use numbering

Output Format:
[term 1, term 2, term 3, term 4, term 5, term 6, term 7,...]

Input:
{reviews}
\end{PromptBox}
\caption{Prompt for generating Single-Sum QA instance.}
\label{fig:prompt_single_summarization}
\end{figure*}

\begin{figure*}[h]
\begin{PromptBox}{Prompt for generating QA instance of Pair-Comp}
You are a professional academic paper analysis expert. Please generate several sets of high-quality comparison questions based on the content of one or several provided papers. Each set of Q&A must be strictly based on the text content, without introducing external knowledge.

Input:
- Paper Content: {paper_content}
- Paper Title: {paper_title}

Task requirements are as follows:
Identify two different methods / techniques / models / methodologies. In the provided text, find two methods, techniques, systems, models, or research paths that are explicitly described or contrasted (e.g., "Government Agencies vs. Industry and AGI Labs" or "RAP vs. MCTS"). Then generate a single, specific, and directly comparative question. The question should focus on one of the following aspects: performance differences, efficiency comparison, applicability scenarios, limitations or risks, basic principles or design philosophy, or governance roles / functional divisions. The question must be expressed in a clear interrogative sentence and should encourage the revelation of differences, trade-offs, or similarities.

Provide a concise keyword-style answer. The answer should be a structured list of 3-10 keywords, formatted as:
[A: keyword1, keyword2, ...] [B: keyword1, keyword2, ...]
Keywords should be concise and accurate, reflecting the core functions, responsibilities, mechanisms, or characteristics in the text. Avoid using complete sentences.

Provide supporting evidence paragraphs. The evidence field needs to contain all relevant sentences from the original text used to support the Q&A, retaining all citation marks for subsequent supplementation of reference content. All questions and answers must closely follow the text content, highlight technical comparisons, and avoid generalization or vague statements.

Example output format (JSON list):
[
  {
    "question": "...",
    "answer": "[A: ...],[B: ...]",
    "evidence": "..."
  }
]

notes:
- All information must originate from the provided text.
- Keywords should be short and powerful, reflecting core concepts.
- Do not generate repetitive or semantically similar questions.
- The complete content should be placed inside; do not summarize or rewrite the evidence sentences, they must be the original text.
- Output in JSON format.
\end{PromptBox}
\caption{Prompt for generating QA instance of Pair-Comp.}
\label{fig:prompt_double_comparison}
\end{figure*}

\clearpage


\newpage

\end{document}